\documentclass[12pt]{article}
\usepackage[margin=1in]{geometry}

\usepackage{amsfonts, amssymb, amsmath, amsthm, mathtools}
\usepackage{graphicx, multicol, verbatim}
\usepackage{cancel}
\usepackage[shortlabels]{enumitem}
\usepackage{array}
\usepackage{latexsym}
\usepackage{geometry}
\usepackage{fancyhdr}
\usepackage{csquotes}
\usepackage[ruled,vlined]{algorithm2e}
\usepackage{multirow}
\usepackage{subcaption}
\usepackage{caption}
\usepackage{mathrsfs}
\usepackage{color}
\usepackage[utf8]{inputenc}

\usepackage{multirow}
\usepackage{subcaption}
\usepackage{tabularx}
\usepackage{booktabs}

\usepackage[most]{tcolorbox}
\usepackage{authblk}
\usepackage{expl3}
\usepackage{hyperref}
\usepackage{cleveref}
\usepackage[round]{natbib}

\usepackage{algorithmic}

\newcommand{\R}{\mathbb R}

\newcommand{\dW}{\mathrm{W}}

\newtheorem{theorem}{Theorem}[section]

\newtheorem{corollary}[theorem]{Corollary}

\newtheorem{definition}[theorem]{Definition}

\newtheorem{problem}[theorem]{Problem}

\definecolor{darkred}{rgb}{1, 0.1, 0.3}
\definecolor{darkgreen}{rgb}{0.5, 0.8, 0.1}
\definecolor{darkpurple}{rgb}{1.0, 0, 1.0}
\definecolor{darkblue}{rgb}{0, 0, 1.0}

\begin{document}

\title{Learning Ultrametric Trees for Optimal Transport Regression}

\author[1]{ Samantha Chen \thanks{\url{sac003@ucsd.edu}}}
\author[2]{ Puoya Tabaghi \thanks{\url{ptabaghi@ucsd.edu}}}
\author[2]{Yusu Wang 
\thanks{\url{yusuwang@ucsd.edu}}}

\affil[1]{Department of Computer Science and Engineering, University of California - San Diego }
\affil[2]{Hal{\i}c{\i}o\u{g}lu Data Science Institute, University of California - San Diego }

\maketitle

\begin{abstract}
Optimal transport provides a metric which quantifies the dissimilarity between probability measures. For measures supported in discrete metric spaces, finding the optimal transport distance has cubic time complexity in the size of the space. However, measures supported on trees admit a closed-form optimal transport that can be computed in linear time. In this paper, we aim to find an optimal tree structure for a given discrete metric space so that the tree-Wasserstein distance approximates the optimal transport distance in the original space. One of our key ideas is to cast the problem in ultrametric spaces. 
This helps us optimize over the space of ultrametric trees --- a mixed-discrete and continuous optimization problem --- via projected gradient decent over the space of ultrametric matrices. During optimization, we project the parameters to the ultrametric space via a hierarchical minimum spanning tree algorithm, equivalent to the closest projection to ultrametrics under the supremum norm. Experimental results on real datasets show that our approach outperforms previous approaches (e.g. Flowtree, Quadtree) in approximating optimal transport distances. Finally, experiments on synthetic data generated on ground truth trees show that our algorithm can accurately uncover the underlying trees.
\end{abstract}

\section{Introduction}
\label{sec:introduction} 
First formulated by Gaspard Monge in 18th-century France, the optimal transport problem is often explained by analogy to the problem of minimizing the time spent transporting coal from mines to factories. More formally, given two distributions and a transportation cost, optimal transport aims to find the lowest-cost way of moving points from the first distribution to the second one. When the transportation cost is a metric, the optimal transport distance is also called the 1-Wasserstein distance~\citep{villani2009optimal}.

Optimal transport is applied in many areas such as machine learning~\citep{solomon2014wasserstein,frogner2015learning,montavon2016wasserstein,kolouri2017optimal,arjovsky2017wasserstein,genevay2016stochastic,lee2018minimax}, statistics~\citep{el2012bayesian,reich2013nonparametric,panaretos2016amplitude}, and computer graphics~\citep{dominitz2009texture,rubner2000earth,rabin2011wasserstein,lavenant2018dynamical,solomon2015convolutional}. Since the optimal transport distance is computationally expensive (cubic in the number of points), several methods have emerged to efficiently approximate optimal transport distance. One of the most popular methods is the Sinkhorn distance, which uses entropic regularization to compute an approximation of optimal transport in \emph{quadratic} time~\citep{cuturi2013sinkhorn}. 


Another approach relies on approximating the original metric with a tree metric~\citep{evans2012phylogenetic,le2019tree2,indyk2003fast,takezawa2021fixed,yamada2022approximating}. While this approach yields a coarser approximation of the optimal transport, it has \emph{linear} time complexity with respect to the size of the metric space. The classic Quadtree is one of the most widely used tree approximations methods. It recursively partitions a metric space into four quadrants to construct internal nodes and represents each element of the original discrete metric space as a leaf node. Quadtree Wasserstein distance approximates the true optimal transport distance with a logarithmic distortion~\citep{indyk2003fast}. Flowtree~\citep{Flowtree} and sliced-tree Wasserstein~ \citep{le2019tree} are Quadtree-based methods designed to improve over the 1-Wasserstein distance approximation of the Quadtree algorithm. A main drawback of Quadtree-based methods is that they require a Euclidean embedding of the original discrete metric space. In contrast, clustertree is a tree-based approximation that does not require a Euclidean embedding of the original discrete metric space~\citep{le2019tree}. Given a \emph{fixed} Quadtree or clustertree topology, \citet{yamada2022approximating} propose a state-of-the-art method based on \emph{optimizing the weights} on the tree to best approximate Wasserstein distances but does not change the topology of the input tree.

Our goal is to find the tree \emph{topology and weight} that closely approximates the Wasserstein distance in the original metric space. To achieve this, we introduce a projected gradient descent procedure over the space of ultrametrics to find a tree that approximates the original Wasserstein distances. As constraining the problem to tree metrics is challenging in general, we instead use ultrametrics --- a subfamily of tree metrics --- as a proxy for tree weights and structure.



{\bf Problem Statement} (\emph{informal}) 
Let $\mathcal{X}$ be a point set in a metric space and $\dW_1(\cdot, \cdot)$ be the optimal transport distance. We want to learn an ultrametric on $\mathcal{X}$ such that ultrametric optimal transport $\dW_{u}$ approximates $\dW_1$. To achieve this, we cast this as a regression problem over ultrametric trees.

The key advantage of optimizing via ultrametrics is that we can project any (semi)metric to the ultrametrics; see \Cref{sec:ultrametric_projection}. This projection allows us to optimize in the space of ultrametrics via a projected gradient descent-type procedure. Our contributions are as follows.

{\bf 1.} We define a quadratic cost function to measure the discrepancy between the true and ultrametric optimal transport distances. This cost does not require the point positions a priori but rather is parameterized by pairwise dissimilarity measures between points, i.e., it \emph{does not} assume that input points are embedded, or even equipped with a metric. 

{\bf 2.} We then propose a projected gradient descent method to perform optimization in ultrametric spaces. The proposed optimization process learns a weighted tree structure. Our proposed method adjusts \textit{both tree weights and structure} throughout training.  This is a novel contribution to existing work on tree-Wasserstein approximation. In previous methods, either the tree structure is fixed~\cite{yamada2022approximating,Flowtree} or it is determined by approximating the discrete metric --- not the Wasserstein distances~\cite{indyk2003fast,le2019tree}. 

{\bf 3.} The learned ultrametric trees provide more accurate optimal transport approximations compared to Flowtree and Quadtree for various real-world datasets. Our method performs slightly worse than the weight-optimized methods~\citep{yamada2022approximating} on real-world distributions {\it with sparse support}; however, for denser synthetic distributions, it outperforms all aforementioned methods. The computational complexity of approximating the optimal transport distance at inference time with our learned ultrametric tree is $O(N)$ similar to other tree approximation methods.

\paragraph{Notation.} For $N \in \mathbb{N}$, we define $[N] = \{1, \ldots, N \}$. We denote the set of nonnegative real numbers as $\mathbb{R}_{+}$. For a vector $x \in \mathbb{R}^d$, we denote its $\ell_p$ norm as $\| x \|_p$. Given a metric space $\mathcal{X}$, we denote the space of probability measures over $\mathcal{X}$ as $\mathcal{P}(\mathcal{X})$. For any $x,y \in \mathbb{R}$, we let $x \land y = \max \{x,y \}$. For matrices $X, Y \in \mathbb{R}^{d_1 \times d_2}$, we let $\langle X , Y \rangle = \mathrm{tr}(X^{\top} Y)$. Let $\mathcal{X}$ to be a finite discrete set. A semimetric over $\mathcal{X}$ is the function $d_s: \mathcal{X} \times \mathcal{X} \rightarrow \R_{+}$ that does not necessarily satisfy the triangle inequality.

\begin{figure*}[t!]
\centering
\includegraphics[scale = 0.55]{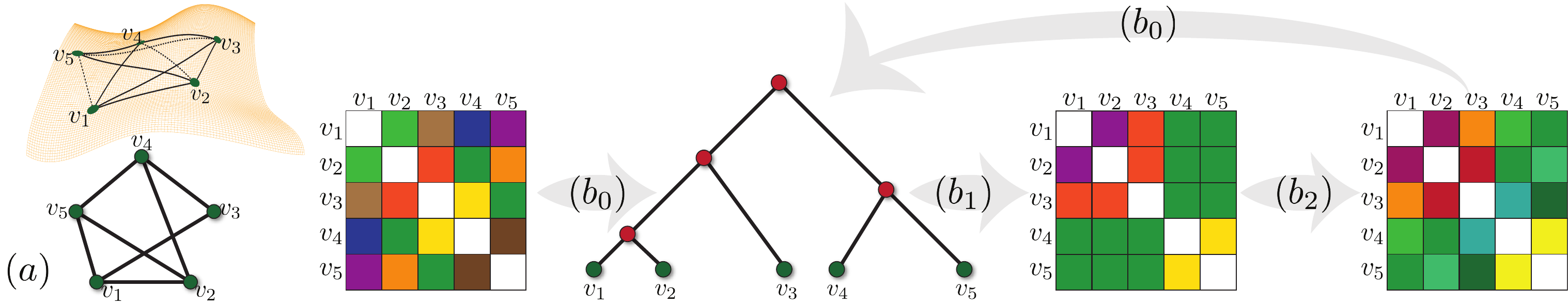}
\caption{ An overall summary of the our projected gradient descent procedure. $(a)$ Measurements are points in a metric space, vertices of a weighted graph, or a distance matrix. $(b_0)$ Hierarchical minimum spanning tree builds an ultrametric tree given a semimetric matrix. $(b_1)$ We compute a distance matrix for the tree leaves. $(b_2)$ We update the distance matrix by applying gradient descent on the optimal transport regression cost.}
\label{fig:procedure}
\end{figure*}

\section{Preliminaries} \label{sec:Wasserstein distance}
\paragraph{Wasserstein Distance.}The Wasserstein distance provides a metric for the space of probability distributions supported on a compact metric space. We focus on the 1-Wasserstein distance (or the optimal transport distance) for discrete probability distributions. For a discrete set $\mathcal{X} = \{ x_n: n \in [N] \}$, we can compute the Wasserstein distance by solving the following linear programming problem:

\begin{equation}
    \dW_1(\mu, \rho) = \min_{ \Pi \in \mathbb{R}_{+}^{N \times N}}   \{\langle \Pi, D \rangle: \Pi 1 = \rho, \Pi^{\top} 1 = \mu\}
\end{equation}
where $D = \big(d(x_{n_1}, x_{n_2}) \big)_{n_1,n_2 \in [N]} \in \mathbb{R}_{+}^{N \times N}$ is the distance matrix, measures $\mu$ and $\rho$ are $N$-dimensional vectors, viz., $1^{\top} \mu = 1, \mu \geq 0$. When $D$ is any arbitrary cost matrix, we can still solve the optimal transport problem. Solving this linear programming problem has a time complexity of $O(N^3 \log{N})$~\citep{pele2009fast}.

\paragraph{Tree Wasserstein Distance.}\label{sec:twd}
Consider a weighted tree $T = (V , E)$ with metric $d_T \in V \times V \rightarrow \mathbb{R}_{+}$. For nodes $v_1, v_2 \in V$, let $P_{v_1, v_2}$ be the unique path between them, and let $\lambda$ be the length measure on $T$ such that $d_T(v_1,v_2)= \lambda( P_{v_1,v_2} )$. We define $T_v$ as the set of nodes contained in the subtree of $T$ rooted at $v_r \in V$, i.e., $T_{v} = \{ v^{\prime} \in V: v \in P_{v_r, v^{\prime}} \}$. Given the metric space $(T, d_T)$ and measures $\mu$ and $\rho \in \mathcal{P}(T)$, \Cref{thm:twd} provides a closed-form expression for the $1$-Wasserstein distance $\dW_{T} (\mu, \rho)$.
\begin{theorem}\label{thm:twd}
Given two measures $\mu,  \rho$ supported on $T = (V,E)$ with metric $d_T$, we have
\begin{equation}\label{eq:tree distance}
    \dW_{T} (\mu, \rho) = \sum_{e \in E} w_e |\mu (T_{v_e} ) - \rho( T_{v_e})|,
\end{equation}
where $w_e$ is the weight of edge $e \in E$, and $v_e$ is the node of $e \in E$ that is farther from the root~\citep{le2019tree}.
\end{theorem}
From \Cref{thm:twd}, we can compute the tree Wasserstein distance using a simple greedy matching algorithm where \emph{mass} from measures $\mu$ and $\rho$ is pushed from child to parent nodes and matched at parent nodes. This involves computing $|\mu (T_{v_e} ) - \rho( T_{v_e})|$ for all nodes $\{ v_e: e \in E\}$. We denote the optimal coupling associated with the tree Wasserstein distance as $\Pi^{T}_{\mu,\rho} \in \Gamma(\mu,\rho)$. \Cref{thm:twd} also provides a natural embedding for probability distributions on a tree to the $\ell_1$ space, as stated in \Cref{cor:l_1_embedding_of_probs}. See Appendix A for discussion on tree Wasserstein. 
\begin{corollary}\label{cor:l_1_embedding_of_probs}
Let $\mathcal{W}_T$ be the set of probability measures defined on a tree $T$. Then, the tree Wasserstein space can be isometrically embedded in the $\ell_1$ space. 
\end{corollary}



\begin{figure*}[th]
    \centering
    \includegraphics[scale=0.24]{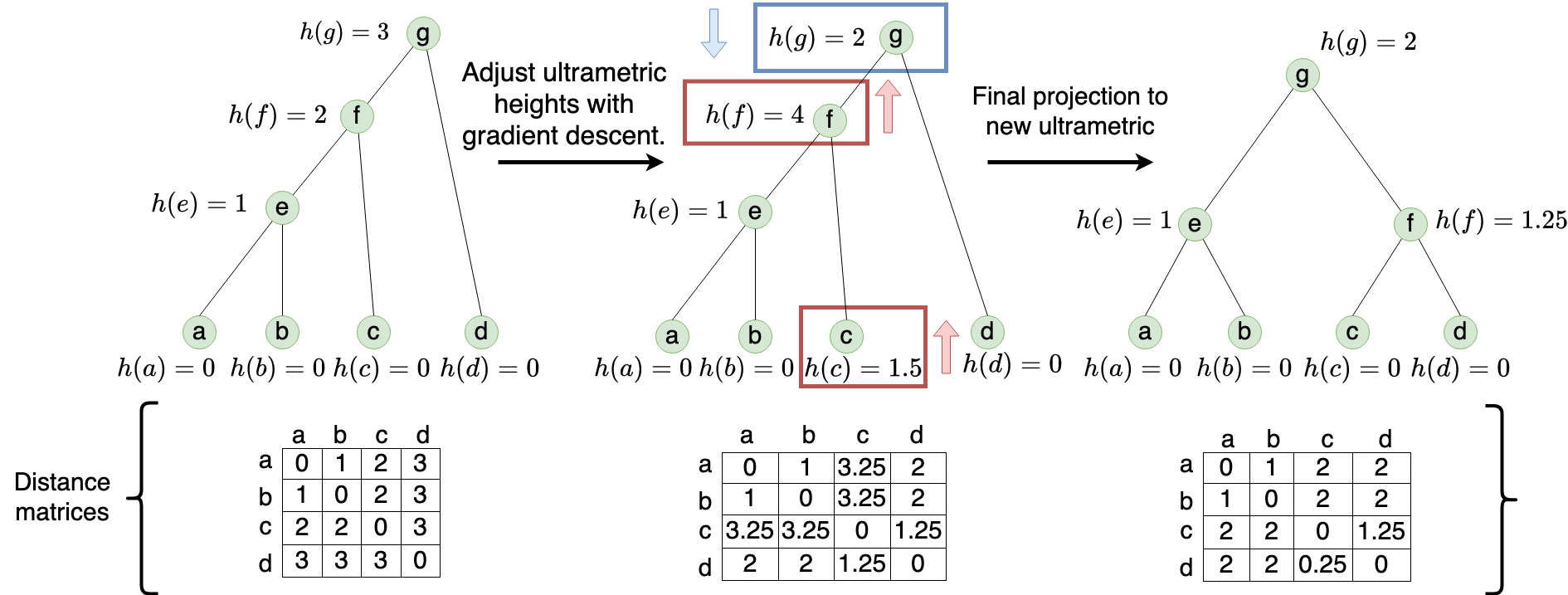}
    \caption{ A stylized example of how the tree structure vary with heights of least common ancestors. Notice that in the second step, the heights of nodes $f$ and $c$ have increased while the height of node $g$ has decreased. Due to these changes in height, the recomputed distance between nodes $c$ and $d$ is smaller than the recomputed distance between $a$ and $b$ so $c$ and $d$ are the first to be clustered in the final projection to the new ultrametric (instead of node $c$ clustering with $a$ and $b$ in the original ultrametric).}
    \label{fig:changing-tree-structure-example}
\end{figure*}


\begin{table*}[h]
\caption{Mean relative error \Big($\frac{|\mathrm{W}_1(\mu, \rho) - \mathrm{W}_T(\mu, \rho)|}{\mathrm{W}_1(\mu, \rho)}$\Big) $\pm$ standard deviation for 1-Wasserstein distance on real-world datasets.} 
\begin{center}
\scriptsize
\begin{tabular}{l|cccccc}
\textbf{Method}  &\textbf{BBCSport} & \textbf{Twitter} &\textbf{RNAseq}  & \textbf{USCA312} & \textbf{USAir97} & \textbf{Belfast} \\
\hline \\
\textbf{Ult. Tree}         & 0.023 $\pm$ 0.010 & 0.062 $\pm$ 0.034 & \textbf{0.041 $\pm$ 0.002}& \textbf{0.282 $\pm$ 0.138} & \textbf{0.043 $\pm$ 0.018} & \textbf{0.368 $\pm$ 0.064}\\
Quadtree          & 0.475 $\pm$ 0.012 & 0.507 $\pm$ 0.019 & 0.929 $\pm$ 0.003 & N/A & N/A & N/A \\
Flowtree          & 0.175 $\pm$ 0.024 & 0.068 $\pm$ 0.036 & 0.079 $\pm$ 0.002 & N/A & N/A & N/A\\
qTWD              & 0.070 $\pm$ 0.020 & 0.045 $\pm$ 0.012 & 0.092 $\pm$ 0.001 & N/A & N/A & N/A \\
cTWD              & 0.028 $\pm$ 0.003 & 0.040$\pm$ 0.029 & 0.044 $\pm$ 0.002 & 0.602 $\pm$ 0.142 & 0.122 $\pm$ 0.008 & 1.089 $\pm$ 0.117\\
Sliced-qTWD       & 0.065 $\pm$ 0.010 & 0.041 $\pm$ 0.016 & 0.091 $\pm$ 0.001 & N/A & N/A &  N/A\\
Sliced-cTWD       & \textbf{0.020 $\pm$ 0.002} & \textbf{0.040 $\pm$ 0.010} & 0.043 $\pm$ 0.002 & 0.674 $\pm$ 0.133 & 0.124 $\pm$ 0.006 & 1.00 $\pm$ 0.133 \\
Sinkhorn, $\lambda = 1.0$ & 0.679 $\pm$ 0.700 & 0.716 $\pm$ 0.075 & 0.981 $\pm$ 0.160 & 0.813 $\pm$ 0.170 & 2.301 $\pm$ 0.144 & 0.713 $\pm$ 0.033
\end{tabular}
\end{center}
\label{tab:real_datasets}
\end{table*}

\section{Optimal Transport Regression in Ultrametric Spaces}
Our goal is to learn a tree metric on a discrete point set such that its optimal transport distance approximates the measured optimal transport distances. We define an optimization problem on ultrametrics as a proxy for tree metrics. 

\begin{definition}
Consider the set $\mathcal{X}$. An ultrametric function $d_u: \mathcal{X} \times \mathcal{X} \rightarrow \mathbb{R}_{+}$ is a metric on $\mathcal{X}$ that also satisfies the strong triangle inequality, i.e.,
\[
    \forall x, y, z \in \mathcal{X}: d_u(x, y) \leq d_u(x, z) \land d_u(y, z) .
\]
\end{definition}
Any compact ultrametric space can be represented by a rooted tree denoted as $T = (v_r, V, E)$, where $v_r$ is the root node and $\mathcal{X}$ is the set of leaves. This representation also includes a height function $h: V \to \mathbb{R}$ with the following property: if $v$ is the lowest common ancestor of the leaves $x$ and $y \in \mathcal{X}$ (denoted $\mathrm{LCA}(x, y)$), then $h(v) = d_u(x, y)$. Furthermore, we induce a weight on the edges of the rooted tree $T$ as follows: given an edge $(v_1, v_2) \in E$ where $v_1$ is closer to the root (or $v_1$ is the parent of $v_2$), we let $w(v_1, v_2) = h(v_1) - h(v_2)$. Then, the weighted tree distance between $x$ and $y$ is related to the heights as follows:
\begin{align}
    d_T(x, y) &= 2\cdot h(\mathrm{LCA}(x, y)) - h(x) - h(y) \label{eq:tree-distance} \\ 
    &= 2\cdot d_u(x, y) - d_u(x, x) - d_u(y, y) \nonumber,
\end{align}
i.e., $d_u(x, y) = \frac{1}{2}d_T(x, y)$. 
For a discrete ultrametric space $(\mathcal{X}, d_u)$, we compute the optimal transport distance as:
\begin{equation*}
    \dW_{u}(\mu, \rho) =  \langle  \Pi^{T}_{\mu, \rho}, D_u \rangle
\end{equation*}
where $\Pi^{T}_{\mu, \rho} \in \Gamma(\mu, \rho)$ is the optimal coupling for the tree $T$ representing the ultrametric space --- constructed via the greedy matching described in \Cref{sec:twd} --- and $D_u = (d_u(x_i, x_j) )_{i,j \in [N]} \in \mathbb{R}_+^{N \times N}$. We now formalize the problem of learning an ultrametric on $\mathcal{X}$ for 1-Wasserstein distance approximation:
\begin{problem} \label{prob:main}
Given an arbitrary discrete set $\mathcal{X}$ endowed with a semimetric, we want to find an ultrametric function $d_u: \mathcal{X} \times \mathcal{X} \to \R_+$  such that for a given set of distributions $\mathcal{S} \subseteq \mathcal{P}(\mathcal{X})$, we minimize the following cost function:
\begin{equation*}
    C(d_u) = \sum_{\mu, \rho \in \mathcal{S}} \Big( \dW(\mu, \rho) - \dW_{u}(\mu, \rho)\Big)^2.
\end{equation*}
\end{problem}
\Cref{prob:main} formulation allows for building an ultrametric on semimetric spaces such as metric spaces and positively weighted undirected graphs. Due to the nonconvex ultrametric constraint, the \Cref{prob:main} is nonconvex  --- in contrast to the optimization problem in \cite{yamada2022approximating}. Since $\mathcal{X}$ is a discrete set of size $N$ (i.e. a metric space with $N$ points), we represent $d_u: \mathcal{X} \times \mathcal{X} \rightarrow \mathbb{R}_{+}$ by $D_u \in \mathbb{UDM}_N$ --- the set of ultrametric distance matrices of size $N$. With slight abuse of notation, we parameterize the cost in \Cref{prob:main} with $D_u$, i.e., $\min_{D_u \in \mathbb{UDM}_N }C(D_u)$. We then minimize the cost using projected gradient descent as follows:
\begin{equation}\label{eq:update_gradient_simplified}
    D_{u}^{(k+1)} =  \mathrm{proj}\big( D_{u}^{(k)} - \alpha \nabla C(D_u^{(k)}) \big),
\end{equation}
where $\alpha$ is the learning rate, $D_{u}^{(k)}$ is the ultrametric matrix at iteration $k$, and $\mathrm{proj}: \mathbb{R}^{N \times N} \rightarrow \mathbb{UDM}_{N}$ projects symmetric matrices to the space of ultrametric matrices of the same size; see \Cref{fig:procedure} for a summary. Entries of $D_{u} \in \mathbb{UDM}_N$ correspond to the pairwise distances between the points in an ultrametric space. The update formula above is a simplified version of the actual procedure; in practice, we do not perform unconstrained optimization over all $N \times N$ matrices. This point will be elaborated in {\bf Computing the Gradient}.

\paragraph{A Projection Operator to Ultrametric Spaces.}  \label{sec:ultrametric_projection} 
Given a semimetric, we use the standard Prim's algorithm to construct an ultrametric tree with leaf nodes in $\mathcal{X}$. We iteratively pick two closest points, e.g., $x_i, x_j \in \mathcal{X}$, and merge them into a new point $x_{(i,j)}$ --- representing an internal node of the tree. The height of this node is equal to the pairwise distance of its children, i.e., $d_s(x_i, x_j)$. We repeat this process until we build a binary ultrametric tree, $T_\mathcal{X}^u$, with leaf vertices in $\mathcal{X}$. This hierarchical minimum spanning tree procedure serves as our projection, $\mathrm{proj}$, from any semimetric matrix $D \in \mathbb{R}^{N \times N}$ to the ultrametric space $\mathbb{UDM}_N$.

\begin{theorem}
    The ultrametric matrix $D_u^* \in  \mathbb{UDM}_N$ which is closest to $D$ under the $\ell_\infty$ norm is $\mathrm{proj}(D) + \frac{1}{2}\|D - \mathrm{proj}(D)\|_\infty$~\citep{chepoi2000approximation}.
\end{theorem}
Since the 1-Wasserstein distance on the ultrametric is \emph{shift-invariant}, we opt to simply use $\mathrm{proj}(D)$. 

\paragraph{Computing the Gradient.} The cost $C(D_u)$ is a quadratic function of $D_u$. At iteration $k$, we fix the tree structure $T^{(k)}$ and compute the gradient of $C(D_u)$ over $\R^{N \times N}$, as follows:
\begin{equation}\label{eq:vanilla_grad}
    \nabla C(D_u^{(k)}) = \sum_{\mu, \rho \in \mathcal{S}} \Big( \langle \Pi^{T^{(k)}}_{\mu, \rho} D_u^{(k)} \rangle -\dW(\mu, \rho)  \Big) \Pi^{T^{(k)}}_{\mu, \rho},
\end{equation}
where $\mathcal{S} \subseteq \mathcal{P}(\mathcal{X})$.
We then adjust the tree structure in the next step --- after projecting the updated distance matrix onto ultrametric matrices.
Additionally, notice that the gradient computed in \eqref{eq:vanilla_grad} implies each element of $D_u$ is an independent variable. 
However, this matrix can only have $2N - 1$ free parameters because it corresponds to an ultrametric tree where leaf nodes with the same least common ancestor (LCA) have the same distance, i.e., the height of their LCA.  Therefore, we first parameterize $D_u$ with $2N-1$ free variables associated with the height of each node in the tree, that is, $(D_u)_{i, j} = \theta_k$ if $\theta_k = h(\mathrm{LCA}(x_i, x_j))$ where $k \in [2N-1]$. In other words, we update both $(D_u)_{i, j}$ and $(D_u)_{i^{\prime}, j^{\prime}}$ in the same way if $\mathrm{LCA}(x_i, x_j) = \mathrm{LCA}(x_{i^{\prime}}, x_{j^{\prime}})$. This is analogous to adjusting height parameters, $\{ \theta_k \}_{k \in [2N-1]}$, in the ultrametric tree associated with $D_u$. After a gradient descent step, $\widehat{D} = D - \nabla_{\theta} C( D_u)$, diagonal elements of $\widehat{D}$ may no longer be zero.  This is because $\widehat{D}_{i, i}$ is associated with the height of leaf node $v_i$ on the tree, and leaf heights may deviate from their typical value of zero after each update. To get a valid distance matrix, we use equation \eqref{eq:tree-distance} and convert new height parameters to ultrametric distances, viz., $\widehat{D}_{i, j} \leftarrow \frac{1}{2} \cdot (2 \cdot \widehat{D}_{i, j} - \widehat{D}_{i, i} - \widehat{D}_{j, j})$ for all $i, j \in [N]$. We then use the updated matrix $\widehat{D}$ as the input for the projection to ultrametric space, $\mathrm{proj}$. The parameterization based on LCA heights and the accompanying post-gradient processing based on equation~\eqref{eq:tree-distance} are the main distinctions to the simplified update rule in equation \eqref{eq:update_gradient_simplified}.




After a gradient descent step, the height of a parent node may become less than its children and the height of leaf nodes may change. This affects the distance matrix $\widehat{D}$ (see equation \ref{eq:tree-distance}) in a way the causes the in the topology of estimated trees during training.
See \Cref{fig:changing-tree-structure-example} for an example of how the tree structure changes with the height of nodes.


\begin{algorithm}[t]
\caption{Ultrametric tree optimization procedure}
\label{alg:final-tree-algorithm}
\textbf{Input}: discrete set $\mathcal{X}$, the distance matrix $D$ for $\mathcal{X}$, learning rate $\alpha$, maximum number of iterations $t_{max}$, training samples $\mathcal{S}$, minimum spanning tree algorithm $\mathrm{MST}$  \\
\textbf{Output}: An ultrametric $d_u$ and associated tree $T_\mathcal{X}^u$
\begin{algorithmic}[1] 
\STATE $D_u^{(0)} = \mathrm{proj}(D)$
\STATE $T_{\mathcal{X}}^{(0)} = \mathrm{MST}(D)$ (simultaneous during $\mathrm{proj}$)
\STATE Let $t=0$.
\WHILE{$t < t_{max}$}
\STATE Compute $C(d_u)$ for training samples $\mathcal{S}$. 
\STATE Given $i, j \in [N]$, $\forall i', j' \in [N]$ such that $\mathrm{LCA}(i, j) = \mathrm{LCA}(i', j')$, $\widehat{D}_{i', j'} \leftarrow (D_u^{(k)} - \alpha \nabla C(D_u^{(k)}))_{i', j'}$
\STATE $\forall i, j  \in [N]$,  $\widehat{D}_{i, j} \leftarrow \frac{1}{2} \cdot (2 \cdot \widehat{D}_{i, j} - \widehat{D}_{i, i} - \widehat{D}_{j, j})$. 
\STATE $D_u^{k + 1} = \mathrm{proj}(\widehat{D})$
\STATE $T_{\mathcal{X}}^{(k + 1)} = \mathrm{MST}(\widehat{D})$ (simultaneous during $\mathrm{proj}$)
\ENDWHILE
\STATE \textbf{return} $D_u^{t_{max}} , T_{\mathcal{X}}^{t_{max}}$
\end{algorithmic}
\end{algorithm}

\paragraph{Runtime Analysis.}
Since we construct the ultrametric (and associated tree) for each input matrix $D$ using the hierarchical minimum spanning tree procedure, the number of nodes in the ultrametric tree is bounded by $O(N)$. Therefore, computing the optimal flow and the Wasserstein distance over the tree takes at most $O(N)$ time --- linear in the number of points. However, computation complexity of the projection algorithm is $O(N^2)$ --- a computational bottleneck for the proposed method. Therefore, the overall complexity of our training is $O(k \cdot N^2)$ where $k$ is the number of gradient descent iterations. This training time is slower than that of \citep{yamada2022approximating}. Nevertheless, for a fixed discrete metric space $\mathcal{X}$, we only need to train the tree once and then we approximate Wasserstein distances in $O(N)$ time, i.e., the same as the inference time for \citep{yamada2022approximating}. 

\paragraph{Why Use Ultrametrics?} We use ultrametrics in \Cref{prob:main} as a proxy for general tree metrics. 
Given some metric $d$, the closest ultrametric $d_u$ provides a 3-approximation for the tree metric to $d$ \cite{agarwala1998approximability}. This is an upper bound on the distortion from the optimal tree metric caused by relaxing the problem to ultrametrics. This fact indicates that optimizing over the space of ultrametrics (as we do in \Cref{prob:main}) is not overly restrictive compared to optimizing over the space of all tree metrics. Furthermore, ultrametrics are widely used in hierarchical clustering applications --- most notably in bioinformatics to model phylogenetic trees. There is also an ultrametric that incurs a $\log n$ distortion with respect to Euclidean distance, i.e. $\|x - y\|_2 \leq d_u(x, y) \leq c \cdot \log n \|x - y\|_2$ \cite{fakcharoenphol2003tight}. Therefore, the ultrametric Wasserstein distance has a $\log n$ distortion compared to the original Euclidean 1-Wasserstein distance i.e. $\mathrm{W}_1(\mu, \rho) \leq \mathrm{W}_u(\mu, \rho) \leq \log n \mathrm{W}_1(\mu, \rho)$. However, this \textit{does not} guarantee that the our projected gradient descent algorithm will achieve this $\log n$ distortion after training.

\section{Experimental Results}
We use PyTorch to implement our tree optimization algorithm, denoted \emph{Ult. Tree}.\footnote{All code is publicly available at github.com/chens5/tree-learning.} We compare the performance of our method with Sinkhorn distance~\citep{cuturi2013sinkhorn}, Flowtree~\citep{Flowtree}, Quadtree~\citep{indyk2003fast}, weight optimized cluster tree Wasserstein distance (cTWD), weight optimized Quadtree Wasserstein distance (qTWD), and their sliced variants~\citep{yamada2022approximating}. All reported results for Sinkhorn distance are computed with the Python Optimal Transport\citep{flamary2021pot} library and with regularization parameter $\lambda = 1.0$. We do not compare the performance of our method to the sliced tree Wasserstein distance of \cite{le2019tree} since sliced-cTWD and sliced-qTWD consistently improve upon the results of \cite{le2019tree}.
We not only compare the approximation accuracy in our experiments but also devise experiments which illustrate the benefits of changing tree structure throughout optimization compared to only learning weights for a fixed tree structure--- as is the case in cTWD and qTWD methods~\citep{yamada2022approximating}. Refer to Appendix C for more details regarding experiments, including dataset generation.

\subsection{1-Wasserstein Approximations}

\paragraph{Real-world Datasets.} We compare 1-Wasserstein distance approximations for the Twitter and BBCSport word datasets~\citep{huang2016supervised} where training data consists of word frequency distributions per document. 
We also use three graph datasets: $(1)$ USCA312, $(2)$ USAir97~\citep{nr} and $(3)$ the Belfast public transit graph~\citep{kujala2018collection}. Finally, we include a high-dimensional RNAseq dataset (publically available from the Allen Institute) which consists of vectors in $\R^{2000}$~\citep{yao2021taxonomy}. Additional details regarding datasets are in Appedix C.
We summarize the error in~\Cref{tab:real_datasets} and the average runtime for each method in~\Cref{tab:computation_time}. Our ultrametric optimization method is slower than cTWD and qTWD in practice (although it has the same theoretical time complexity) because the size of the ultrametric tree generated by MST is larger than cTWD and qTWD.



\begin{table}[t]
\caption{Average runtime for computing $1$-Wasserstein distance of each algorithm based on their CPU implementations. The time complexity of the Quadtree-based methods, e.g., qTWD and sliced-qTWD, are similar to the Quadtree.}
\begin{center}
\begin{tabular}{l|lll}
\textbf{Method}   & \textbf{USCA312} & \textbf{USAir97} & \textbf{Belfast} \\
\hline
\textbf{Ult. Tree}        & 0.00006 & 0.00013 & 0.00944\\
cTWD             & \textbf{0.00004} & \textbf{0.00002} & \textbf{0.00009}\\
Sinkhorn, $\lambda$=1.0 & 0.01450 & 0.01883 & 0.4327\\
OT               & 0.00692 & 0.03819 & 0.5458\\
\hline 
\textbf{Method}  &\textbf{BBCSport} & \textbf{Twitter} & \textbf{RNAseq} \\
\hline 
\textbf{Ult. Tree}         & 0.4204 & 0.1310 & 0.0017\\
Quadtree          & \textbf{0.0001} & \textbf{0.0001} & \textbf{0.0001}\\
Flowtree          & 0.0002 & 0.0001  & 0.0001\\
cTWD              & 0.0004 & 0.0002 & 0.0002\\
Sinkhorn, $\lambda$=1.0 & 2.0851 & 0.3601 & 7.612\\
OT                & 1.6001 & 0.3820 & 0.570\\
\end{tabular}
\end{center}

\label{tab:computation_time}
\end{table}

\begin{table*}[t]
    \centering
    
\label{tab:synthetic_random}
\caption{Mean relative error for approximating $1$-Wasserstein distance on a synthetic dataset of $100$ randomly sampled from the uniform distribution over $[-10, 10]^d$. The training data consists of $200$ randomly generated distribution pairs.}
\begin{center}
\scriptsize
\begin{tabular}{l|cccccc}
\textbf{Method}  &\textbf{dim=2} & \textbf{dim=5} & \textbf{dim=8} &\textbf{dim=11} &\textbf{dim=14} &\textbf{dim=17} \\
\hline \\
\textbf{Ult. Tree}         & \textbf{0.104 $\pm$ 0.074} & \textbf{0.029 $\pm$ 0.023} & \textbf{0.016 $\pm$  0.013} & \textbf{0.015 $\pm$ 0.011} & \textbf{0.012 $\pm$ 0.009} & \textbf{0.011 $\pm$ 0.007} \\
Flowtree          & 0.522 $\pm$ 0.145 & 0.559 $\pm$ 0.067 & 0.472 $\pm$ 0.053 & 0.430 $\pm$ 0.039 & 0.378 $\pm$ 0.030 & 0.348 $\pm$ 0.029 \\
Quadtree          & 5.336 $\pm$ 1.046 & 2.981 $\pm$ 0.275 & 1.852 $\pm$ 0.158 & 1.310 $\pm$ 0.093 & 0.823 $\pm$ 0.036 & 0.624 $\pm$ 0.036 \\
qTWD              & 0.557 $\pm$ 0.163 & 0.371 $\pm$ 0.056 & 0.297 $\pm$ 0.037 & 0.267$\pm$ 0.031 & 0.220 $\pm$ 0.022 & 0.219 $\pm$ 0.020 \\
cTWD              & 0.563 $\pm$ 0.177 & 0.371 $\pm$ 0.056 & 0.297 $\pm$ 0.037 & 0.267 $\pm$ 0.032 & 0.221 $\pm$ 0.022 & 0.219 $\pm$ 0.020\\
Sliced-qTWD       &  0.567 $\pm$ 0.132 & 0.387 $\pm$ 0.043 & 0.288 $\pm$ 0.026 & 0.260 $\pm$ 0.022 & 0.218 $\pm$ 0.016  & 0.206 $\pm$ 0.014\\
Sliced-cTWD       & 0.564 $\pm$ 0.138 & 0.387 $\pm$ 0.043 & 0.288 $\pm$ 0.026 & 0.261 $\pm$ 0.022 & 0.219 $\pm$ 0.016 & 0.206 $\pm$ 0.015\\
Sinkhorn, $\lambda = 1.0$ & 0.587 $\pm$  0.257 & 0.071 $\pm$ 0.053 & 0.063 $\pm$ 0.046 & 0.074 $\pm$ 0.062 & 0.068 $\pm$ 0.049 & 0.076 $\pm$ 0.057
\end{tabular}
\end{center}

\end{table*}

We outperform all other methods for the three graph datasets. However, for the two word datasets, Twitter and BBCSport, our error is worse compared to cTWD, qTWD, and their sliced variants. In the word datasets, word frequency distributions generated from documents have limited supports. It seems that this sparsity hinders the proper optimization of our projected gradient descent method as the constrained support of pairs of sparse distributions restricts each step of our optimization to a small portion of the tree. This is not an issue for graph datasets, since we use randomly generated full support distributions for training. 
In all cases, we outperform the distribution-agnostic methods like Quadtree and Flowtree. 

Quadtree approximations \emph{require} embeddings in a Euclidean space. However, for graph datasets, there are no faithful embbeddings for the nodes in a Euclidean space which preserves the semimetric information (from the edge weights). Therefore, we do not conduct experiments with Quadtree-based methods on graph datasets. In contrast, our proposed method, (sliced-)cTWD, and the Sinkhorn algorithms do not rely on the existence of such embeddings.

\paragraph{Synthetic Datasets.}
To compare the performance of our ultrametric tree optimization with all aforementioned algorithms, we generate $100$ random point sets from the uniform distribution in a $d$-dimensional hypercube, e.g., $[-10, 10]^d$, where $d \in \{2, 5, 8, 11, 17\}$, and select $200$ full-support distribution pairs. In \Cref{tab:synthetic_random}, we summarize the mean relative error for approximating the $1$-Wasserstein distances.


Across all dimensions, our learned ultrametric method yields better approximations compared to all other methods by orders of magnitude. 
Furthermore, as our learned ultrametric optimization procedure does not depend on the embedding of the points themselves and only depends on the distance matrix, the variance of error does not vary with the points' dimension. The approximation bounds for Flowtree and Quadtree depends on the dimension and varies (in average) as the dimension changes. Moreover, our ultrametric tree approximation method is faster than the Sinkhorn algorithm (with $\lambda = 1.0$) while yielding a better quality optimal transport approximation.

\subsection{Changing Tree Structure}
One of the main advantages of our method, compared to qTWD and cTWD, is its capability to alter tree topology during training. 
In what follows, we show the utility of this property in both approximating 1-Wasserstein distance and recovering hidden tree metrics. In these experiments, given a tree (metric space) $\mathcal{X}$, we want to approximate 1-Wasserstein distance between distributions on its leaves. We purposefully initialize both our ultrametric learning procedure and cTWD with an incorrect tree topology. 


\begin{figure*}[t]
\includegraphics[width=\linewidth]{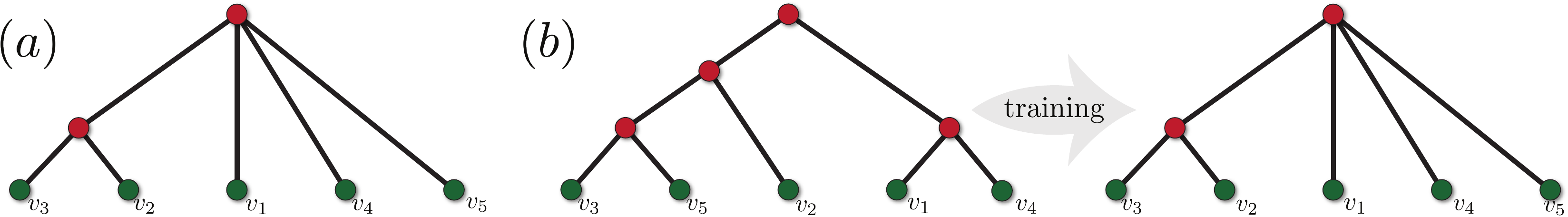}
\caption{ $(a):$ A randomly generated tree $T$ with unit edge weights and leaf distance matrix $D_T$. $(b):$ We initialize the training process with noise-contaminated distance matrix, $\widetilde{D_T}$, and determine the initial tree topology on the left. We then train on true optimal transport distances and correctly recover the original tree structure on the right. }
\label{fig:tree_comparison}
\end{figure*} 

Let us begin with an illustrative example. Consider $\mathcal{X} = \{x_0, x_1, x_2, x_3\}$ with the following distance matrix 
\[
    D = \begin{pmatrix}0 & 4 & 4 & 2\\
                       4 & 0 & 2 & 4\\
                       4 & 2 & 0 & 4\\
                       2 & 4 & 4 & 0 
        \end{pmatrix} \in \mathbb{R}_+^{4 \times 4}
\]
This is the distance matrix for the metric space of a binary ultrametric tree $T$ with first-level clusters $\{x_0, x_4\}$ and $\{x_1, x_2\}$. 
The goal is to approximate Wasserstein distances in the original space $\mathcal{X}$. We first generate $200$ pairs of distributions anc compute their Wasserstein distances in $\mathcal{X}$ (with respect to $D$). We optimize the weights of cTWD with respect $D$ and train our ultrametric tree algorithm.

Now suppose we initialize both algorithms, our ultrametric optimization and cTWD, with the following adversarial distance matrix $D^{\prime}$: 
\[
    D^{\prime} = \begin{pmatrix}0 & 2 & 4 & 4\\
                       2 & 0 & 4 & 4\\
                       4 & 4 & 0 & 2\\
                       4 & 4 & 2 & 0 
        \end{pmatrix} \in \mathbb{R}_+^{4 \times 4},
\]
where $D^{\prime}$ is associated with the binary tree $T^{\prime}$ which first-level clusters $\{x_0, x_1\}$ and $\{x_2, x_3\}$. Given $D'$ as the input distance matrix, $T^{\prime} \neq T$ is the initial tree structure for both algorithms. In the case of cTWD, the tree remains \emph{fixed} as $T^{\prime}$.
For cTWD, after optimizing the weights of $T^{\prime}$, the mean relative error between Wasserstein distance of pairs of distributions on the $T$ (ground truth) and their approximations with cTWD is $0.193 \pm 0.101$. 
On the other hand, our ultrametric tree optimization algorithm changes the tree topology throughout the training, and the final output tree metric is the same as $D$. Therefore, the learned ultrametric tree fully recovers the target metric space --- even given an adversarial start --- and perfectly computes Wasserstein distances. Furthermore, the final estimate tree topology, from our algorithm, is indeed $T$.

We extend this example by taking $100$ random matrices (elements are i.i.d. from uniform distribution on $[10, 20]$) as incorrect initializations for cTWD and our ultrametric tree optimization algorithms.
For \emph{all $100$} random trials, our ultrametric tree optimization algorithms correctly recovers the original tree weights and its topology. Therefore, the learned ultrametric tree Wasserstein distances are exactly the same as the measured distances on $\mathcal{X}$; thereby giving the zero error. In contrast, mean $\pm$ standard deviation of the error for the Wasserstein distance in $\mathcal{X}$ for cTWD is $0.178 \pm 0.103$. These examples demonstrate the significance of correctly estimating the tree topology on approximating the Wasserstein distances, as cTWD incurs error when initialized with incorrect tree structures. 
More importantly, this illustrates the efficacy of our learned ultrametric tree in recovering the precise tree topology, thus enabling accurate computation of Wasserstein distances. 
 \begin{table}[t]
 \caption{Tree metric error and Wasserstein error for the learned ultrametric tree and the skip-MST method given an initial tree metric perturbed by Gaussian noise. }
    \centering
    \begin{tabular}{l|c|cc}
        \textbf{Method} & $\sigma$ &  $\mathrm{dist}(\widehat{T}, T)$ & \textbf{$\mathrm{W}_1$ error}\\
         \hline
         \textbf{Ult. tree} & \multirow{3}*{5}&  \textbf{0.559 $\pm$ 0.218} & \textbf{0.262 $\pm$ 0.102}\\
       skip-MST & &   1.043 $\pm$ 0.323 & 0.279 $\pm$ 0.102\\
       cTWD & & 1.096 $\pm$ 0.281 & 0.285 $\pm$ 0.212 \\
       \hline
        \textbf{Ult. tree} & \multirow{3}*{4}&  \textbf{0.536 $\pm$ 0.183} & \textbf{0.275 $\pm$ 0.257}\\
      skip-MST &   & 0.992 $\pm$ 0.357 & 0.286 $\pm$ 0.109\\
      cTWD & & 1.084 $\pm$ 0.315 & 0.295 $\pm$ 0.113\\
       \hline
       \textbf{Ult. tree} & \multirow{3}*{3}&  \textbf{0.528 $\pm$ 0.213} & \textbf{0.265 $\pm$ 0.092}\\
       skip-MST & &  0.983 $\pm$ 0.268 & 0.281 $\pm$ 0.098\\
       cTWD & & 1.175 $\pm$ 0.388 & 0.270 $\pm$ 0.118\\
       \hline
       \textbf{Ult. tree} & \multirow{3}*{2}&  \textbf{0.506 $\pm$ 0.179} & \textbf{0.268 $\pm$ 0.095}\\
       skip-MST &  & 0.815 $\pm$ 0.280 & 0.283 $\pm$ 0.108\\
       cTWD & & 1.226 $\pm$ 0.358 & 0.293 $\pm$ 0.104
    \end{tabular}
    \label{tab:tree metric reconstruction}
    
\end{table}

\paragraph{Learning Topologies of Random Trees.}
In what follows, we design an experiment to show the impact of changing tree topology learning Wasserstein distances and highlight the importance of changing tree topology from \textit{iteration to iteration} using our minimum spanning tree procedure.

In this experiment, we adjust our ultrametric optimization method to only learn the weights of a fixed tree structure and change tree structure once at the last iteration, resulting in a more efficient procedure by avoiding the $O(N^2)$ complexity of projecting to the ultrametrics at each iteration. We initialize the tree topology using the hierarchical minimum spanning tree algorithm. Throughout training, we update the entries of the distance matrix via gradient descent without projecting onto the ultrametric space. At the end of training, we project the resulting distance matrix to the ultrametric space. This is what we call the \emph{skip-MST method} as we avoid the minimum spanning tree procedure throughout training. 

To compare cTWD, the skip-MST and the regular ultrametric optimization methods, we randomly generate $100$ weighted tree metrics, from random trees with unit weights and $20$ to $40$ nodes. The distance matrix between leaf nodes is then perturbed by a symmetric noise matrix with i.i.d. elements from $\mathcal{N}(0, 2\sigma^2)$. We use this noise-contaminated distance matrix to determine the initial tree topology for all methods. We then synthetically generate probability distributions over the leaves of the underlying tree metrics and compute their exact optimal transport distances. 
We quantify the quality of the estimated tree with the following metric:
\begin{equation*}
    \mathrm{dist}( \widehat{T}, T) = \frac{2}{N(N-1)} \| D_{\widehat{T}} - D_T\|_F,
\end{equation*}
where $T$ is the original tree, $D_T$ is the noiseless distance matrix, and $D_{\widehat{T}}$  is the distance matrix for estimated tree $\widehat{T}$. We report the results in \Cref{tab:tree metric reconstruction}. 

For each value of $\sigma \in \{ 2, 3, 4,5 \}$, the difference between the final and true tree metrics for our ultrametric optimization algorithm is lower than that of both skip-MST and cTWD. This shows that ultrametric optimization adjusts the tree topology to correct for the inaccurate initial topology throughout training. Additionally, an incorrect tree topology has a detrimental effect on the estimated tree metric and optimal transport distance approximation as our ultrametric tree optimization algorithm shows between 4\% and 6\% improvement on Wasserstein error compared to the skip-MST and between 3\% and 10\% improvement compared to cTWD. Furthermore, it is noteworthy that, in most cases, skip-MST manages to achieve a lower error compared to cTWD even though, during training, it uses a fixed initial tree topology.
In \Cref{fig:tree_comparison}, we illustrate an instance of a simple tree and how our ultrametric optimization procedure recovers the underlying tree topology from a bad initialization.  

\section{Conclusion}
We present a novel contribution to the existing lines of work on linear-time optimal transport approximations. We propose a projected gradient descent algorithm that minimizes an optimal transport regression cost and relies on performing optimization in ultrametric space as a proxy for learning the optimal tree metric. The learned ultrametric tree offers an improved approximation for optimal transport distance compared to state-of-the-art tree Wasserstein approximations methods on synthetic datasets and provides a valuable way to recover tree topology. We acknowledge that the projection to ultrametric space is a computational bottleneck for our algorithm. An interesting future direction could be to find a fast projection to ultrametric space. Additionally, optimal transport on trees corresponds to an $L_1$ embedding so our learned tree can be used in locality sensitive hashing data structures for fast nearest neighbor queries. Finally, as our method adjusts tree topology, it is useful in applications where the data has some unknown underlying tree topology.

\section*{Acknowledgements}
The authors would like to thank all anonymous reviewers for their valuable feedback. This work is partially supported by the NSF under grants CCF-2112665, CCF-2217033, and CCF-2310411.

\bibliography{main}

\newpage
\appendix
\onecolumn

\section{Tree approximations of 1-Wasserstein distance}\label{sec:approx_1_wass}
In this section, we overview several methods that build tree metrics on a Euclidean space to approximate the $1$-Wasserstein distance: quadtree, Flowtree, and tree sliced methods.
\subsection{Quadtree}\label{sec:quadtree}
Let $\mathcal{X}$ be a set of $N$ points in $\R^d$ endowed with the $\ell_2$ distance. Without loss of generality, let us have the following: 
\[ 
    \min_{\substack{x_1, x_2 \in \mathcal{X} \\ x_1 \neq x_2}} \| x_1 - x_2\|_2 = 1, \ \max_{x_1, x_2 \in \mathcal{X}} \| x_1 - x_2\|_2 = \Delta,
\]
where $\Delta$ is the \emph{spread} of $\mathcal{X}$ --- the ratio of the diameter of $\mathcal{X}$ over the minimum pairwise distance. 
Quadtree is construted by recursively splitting randomly shifted hypercubes along each dimension until a sub-hypercube contains exactly one point. 
This process has a time complexity of $O(|\mathcal{X}| \log (d\Delta))$. \Cref{thm:quadtree} quantifies an upper bound for the expected approximation error of this algorithm. 

\begin{theorem}\label{thm:quadtree}
Given a randomly shifted quadtree over $\mathcal{X}$, we have the following approximation bound for the optimal transport distance:
\begin{equation*}
    O(1) \leq \frac{\mathbb{E}[\dW_{QT}(\mu, \rho)]}{\dW_1(\mu, \rho)} \leq O(\log d\Delta) ,
\end{equation*}
where $\dW_1(\mu, \rho)$ is the $1$-Wasserstein distance between distinct measures $\mu, \rho \in \mathcal{P}(\mathcal{X})$ \citep{indyk2003fast}.
\end{theorem}
We can compute the quadtree Wasserstein distance in linear time, as follows
\begin{equation}
    \dW_{QT} (\mu, \rho) = \Delta \sum_{e \in E} 2^{-l_e} |\mu (T_{v_e} ) - \rho( T_{v_e})|,
\end{equation}
 where $l_e$ is the side length of the representative sub-hypercube.


\subsection{Flowtree and sliced-tree Wasserstein distances}
Flowtree \citep{Flowtree} is built upon the quadtree approach to approximate $1$-Wasserstein distances with an improved accuracy. This algorithm uses the optimal matching provided by the quadtree (the greedy matching algorithm on the tree). However, instead of using the inexact tree distances, it uses the Euclidean distance between the matched points (i.e. the optimal coupling on the tree). Flowtree has the same theoretical approximation guarantees as quadtree but in practice, provides more accurate results.

\emph{Sliced-tree Wasserstein distance}, proposed in \citep{le2019tree}, is another way to improve the quadtree approximation quality. Instead of approximating 1-Wasserstein distance with one randomly shifted quadtree, sliced-tree Wasserstein distance is the average of tree Wasserstein distances computed on several different random quadtrees. More recently, \citep{yamada2022approximating} proposed an optimization procedure to optimize weights over a fixed tree topology to better approximate 1-Wasserstein distances.

\begin{table*}[b]
\centering
\caption{Mean relative error for approximating $1$-Wasserstein distance on a synthetic dataset sampled from $d$-dimensional white Gaussian distribution. The training data consists of $200$ randomly generated pairs of distributions.}
\scriptsize

\label{tab:synthetic_gaussian}

\begin{center}
\begin{tabular}{l|cccccc}
\textbf{Method}  &\textbf{dim=2} & \textbf{dim=5} & \textbf{dim=8} &\textbf{dim=11} &\textbf{dim=14} &\textbf{dim=17} \\
\hline \\
\textbf{Ult. Tree}         & \textbf{0.109 $\pm$ 0.083} & \textbf{0.031 $\pm$ 0.023} & \textbf{0.018 $\pm$ 0.013} & \textbf{0.015 $\pm$ 0.012} & \textbf{0.016 $\pm$ 0.012} & \textbf{0.012 $\pm$ 0.009}\\
Flowtree          & 0.570 $\pm$ 0.169 & 0.462 $\pm$ 0.068 & 0.406 $\pm$ 0.037 & 0.347 $\pm$ 0.035 & 0.334 $\pm$ 0.028 & 0.298 $\pm$ 0.024\\
Quadtree          & 5.334 $\pm$ 1.263 & 3.217 $\pm$ 0.462 & 2.405 $\pm$ 0.363 & 3.160 $\pm$ 0.283 & 2.555 $\pm$ 0.184 & 2.024 $\pm$ 0.158\\
qTWD              & 0.585 $\pm$ 0.127 & 0.358 $\pm$ 0.036 & 0.297 $\pm$ 0.025 & 0.245 $\pm$ 0.018 & 0.214 $\pm$ 0.015 & 0.188 $\pm$ 0.011 \\
cTWD              & 0.607 $\pm$ 0.120 & 0.358 $\pm$ 0.036  & 0.297 $\pm$ 0.025 & 0.245 $\pm$ 0.019 & 0.213 $\pm$ 0.016 &  0.188 $\pm$ 0.011\\
Sliced-qTWD       & 0.604 $\pm$ 0.125 & 0.358 $\pm$ 0.036 & 0.297 $\pm$ 0.025 & 0.245 $\pm$ 0.018 & 0.213 $\pm$ 0.015 &  0.188 $\pm$ 0.011\\
Sliced-cTWD       & 0.555 $\pm$ 0.134 & 0.350 $\pm$ 0.036 & 0.297 $\pm$ 0.025 & 0.246 $\pm$ 0.018 & 0.213 $\pm$ 0.016 & 0.188 $\pm$ 0.011 \\
Sinkhorn, $\lambda = 1.0$ & 4.881 $\pm$ 1.036 & 2.633 $\pm$ 0.325 & 1.952 $\pm$ 0.276 & 1.432 $\pm$ 0.197 & 0.967 $\pm$ 0.152 & 0.707 $\pm$ 0.151
\end{tabular}
\end{center}

\end{table*}


\section{Trees and ultrametrics}
\label{appendix:additional-ultrametric}
In this section, we will further describe both how to construct the tree associated with a discrete ultrametric space. 
First, we will provide more context as to how to identify a discrete ultrametric space $\mathcal{X}$ with a rooted tree $T_\mathcal{X} = (v_\mathcal{X}, V, E)$ (where $v_\mathcal{X}$ is the root node, $V$ are the vertices, and $E$ are edges) along with a height function $h: V \to \R$ on the nodes of $T_\mathcal{X}$. Note that the leaves of $T$ represent the elements of $\mathcal{X}$. 
To show this, let $(\mathcal{X},d_u)$ be a finite ultrametric space with $|\mathcal{X}| \geq 2$. We can define the diametrical graph  $G_{\mathcal{X}} = (V,E)$ with vertex set $V = \mathcal{X}$ and
\begin{equation}\label{eq:diametrical_graph}
    \forall v_1, v_2 \in V: \{v_1,v_2\} \in E  \Leftrightarrow d_u(v_1,v_2) = h(\mathcal{X}),
\end{equation}
where $h(\mathcal{X})=\mathrm{diam}\big( \mathcal{X} \big)$ is the height function.
The process detailed in \eqref{eq:diametrical_graph}, partitions the vertex set $\mathcal{X}$ into the $k$ first-level node sets, e.g., $\mathcal{X}_1, \ldots, \mathcal{X}_k$ and naturally gives rise to the construction of an ultrametric tree $T_{\mathcal{X}}$. We let $v_{\mathcal{X}}$ be the root vertex of $T_{\mathcal{X}}$. Then, we let $v_{\mathcal{X}_1}, \ldots, v_{\mathcal{X}_k}$ be the children of $v_{\mathcal{X}}$. We repeat the process in \eqref{eq:diametrical_graph} --- on $\mathcal{X}_1, \ldots, \mathcal{X}_k$ --- to determine the next level children. We stop the process until we reach a tree with $|\mathcal{X}|$ leaves. This construction ensures that the nodes in a subtree are equidistant --- equivalent to the height of their least common ancestor.

\section{Additional experimental results}
\label{sec:additional_experiments}
First, we will further describe the experiment setup and datasets used for our ultrametric optimization procedure. For Twitter and BBCSport, we mirror the experimental setup of \cite{huang2016supervised}. Twitter consists of 6344 word embedding vectors in $\R^{300}$ (i.e. the initial input distance matrix $D$ for Twitter is in $\R^{6344 \times 6344}$)  and BBCSport consists of 13243 word embedding vectors in $\R^{300}$. For each word dataset, training data consists of word frequency distributions per document. For Twitter, we first split 3108 probability distributions (generated from word frequencies of documents) into 2176 train distributions and 932 test distributions. The train dataset for the tree consists of all pairs of train distributions and their pairwise Wasserstein distance (i.e. the training dataset $\mathcal{S}$ is defined as $\mathcal{S} = \{(\mu_i, \mu_j, \mathrm{W}_1(\mu_i, \mu_j)) : i, j \in [2176]\}$). Subsequently, the test dataset for the tree is all pairs of test distributions and their pairwise Wasserstein distance. We construct the train and test datasets for BBCSport in the same way, except there are 
737 total distributions (documents), 517 train distributions, and 220 test distributions. 
For the graph datasets, we use: $(1)$ USCA312 which consists of $312$ cities and their pairwise road distances, $(2)$ USAir97~\citep{nr} which consists of $332$ airports and their flight frequencies and $(3)$ the Belfast public transit graph~\citep{kujala2018collection} which consists of $1917$ stops and transit times.
For all graph datasets, we generate 200 random train distributions and construct a train dataset consisting of all pairs of train distributions and their pairwise Wasserstein distance. We use 60 test distributions and generate the test dataset in the same way as above. 
Finally, we also include a high-dimensional RNAseq dataset (publically available from the Allen Institute) which consists of 4360 vectors in $\R^{2000}$. We trained on 20,000 pairs of random distributions and tested on 200 pairs of random distributions.
For all synthetic datasets, we construct the train dataset by taking 200 random generate pairs of distributions and their Wasserstein distance and test our final tree on 50 pairs of distributiosn and their corresponding Wasserstein distance. 

\subsection{Training time and convergence}
\begin{figure*}[t]
    \begin{subfigure}{0.30\textwidth}
    \centering
        \includegraphics[width=\textwidth]{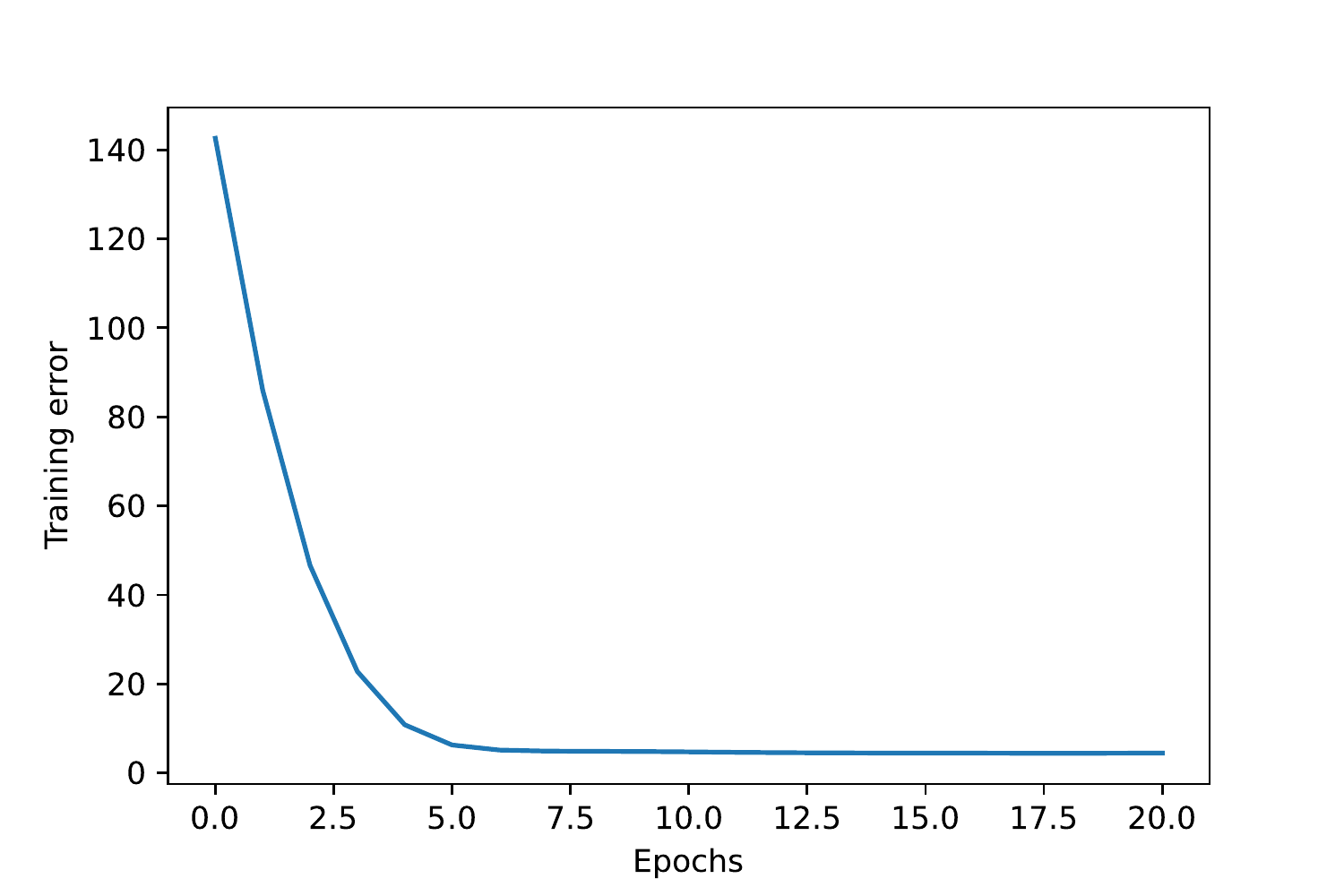}
        \caption{Twitter}
    \end{subfigure}
    \hfill
    \begin{subfigure}{0.30\textwidth}
    \centering
        \includegraphics[width=\textwidth]{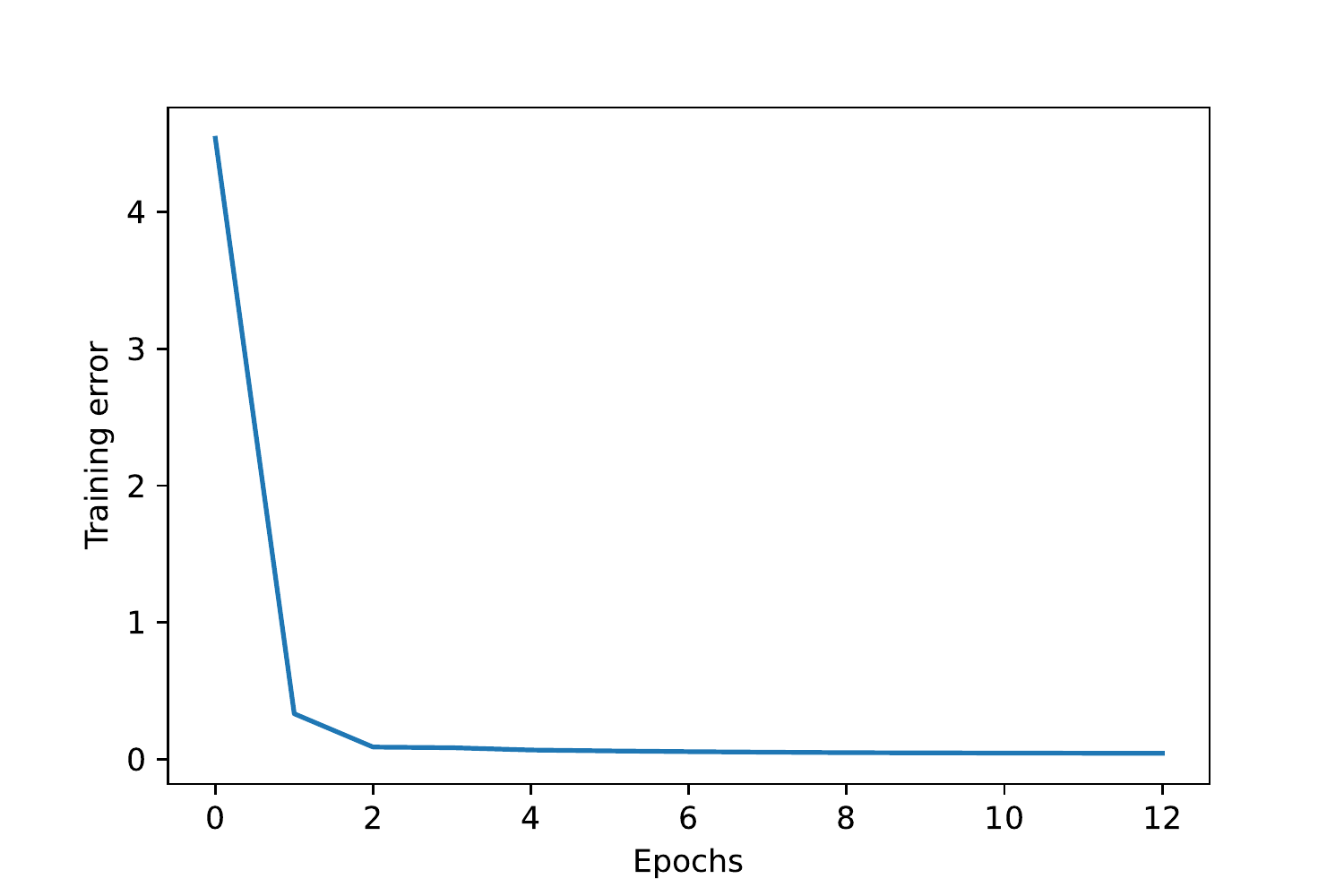}
        \caption{BBCSport}
    \end{subfigure}
    \hfill
    \begin{subfigure}{0.30\textwidth}
    \centering
        \includegraphics[width=\textwidth]{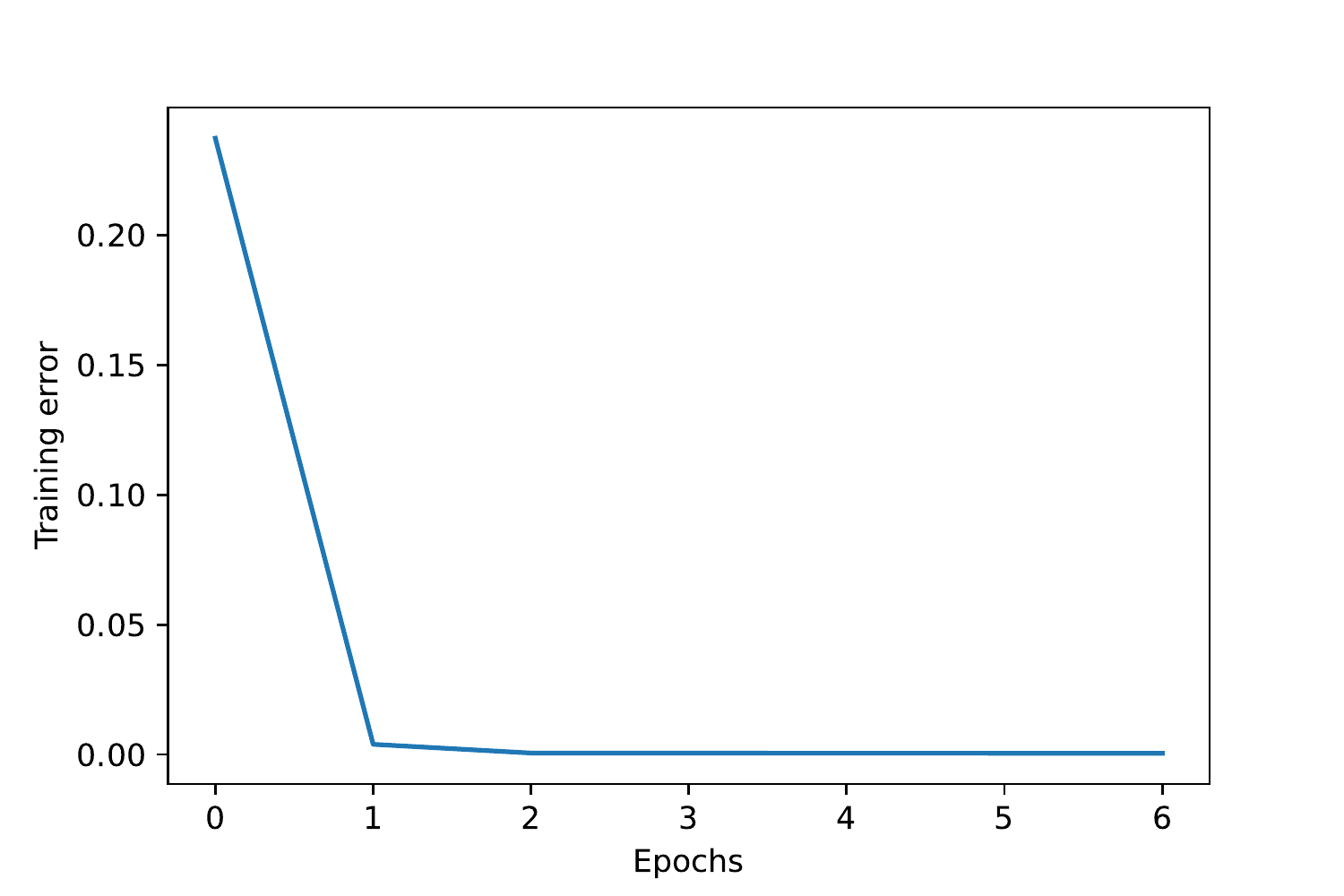}
        \caption{USAir97}
    \end{subfigure}
    \vfill
    \begin{subfigure}{0.30\textwidth}
    \centering
        \includegraphics[width=\textwidth]{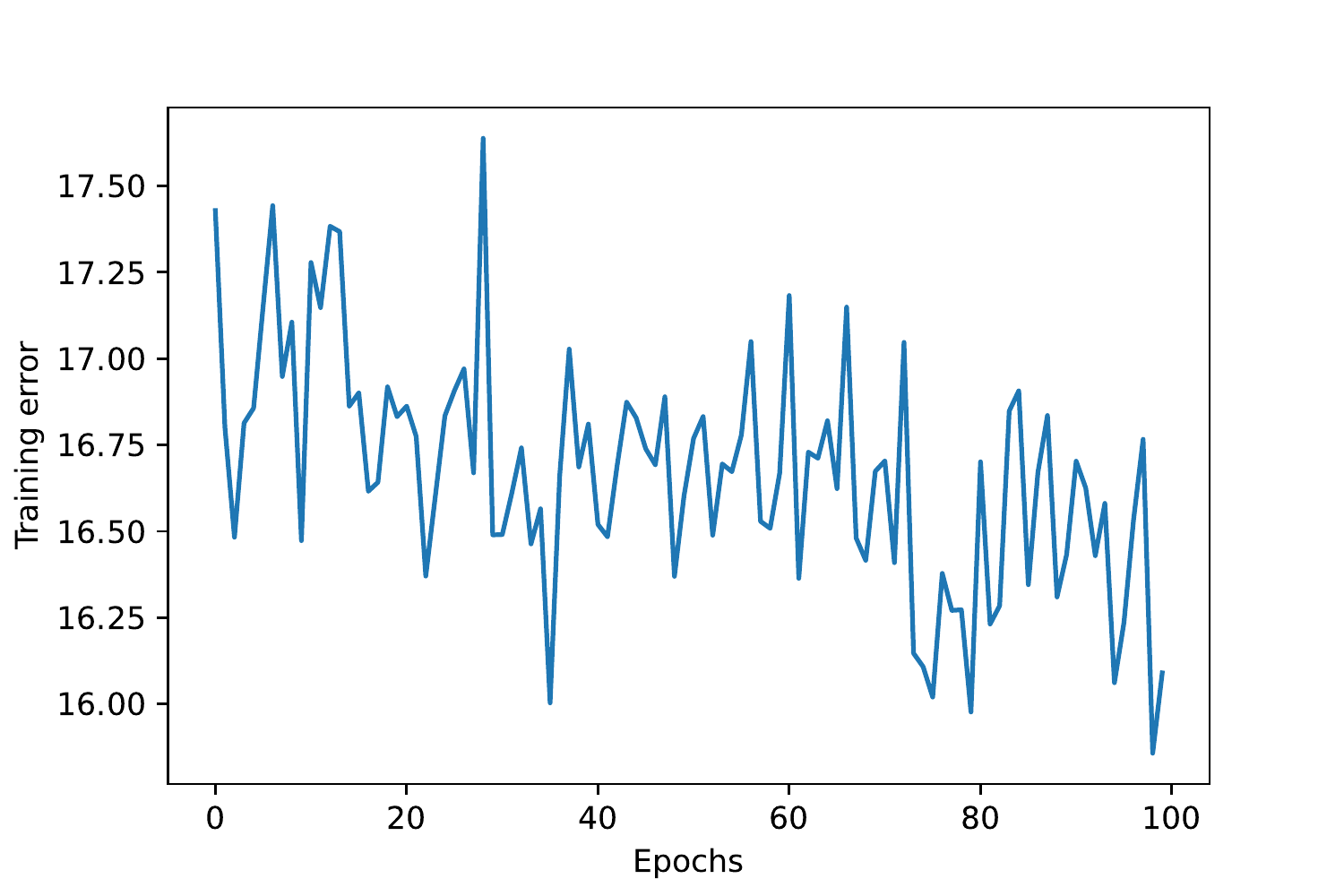}
        \caption{Belfast}
    \end{subfigure}
    \hfill
    \begin{subfigure}{0.30\textwidth}
    \centering
        \includegraphics[width=\textwidth]{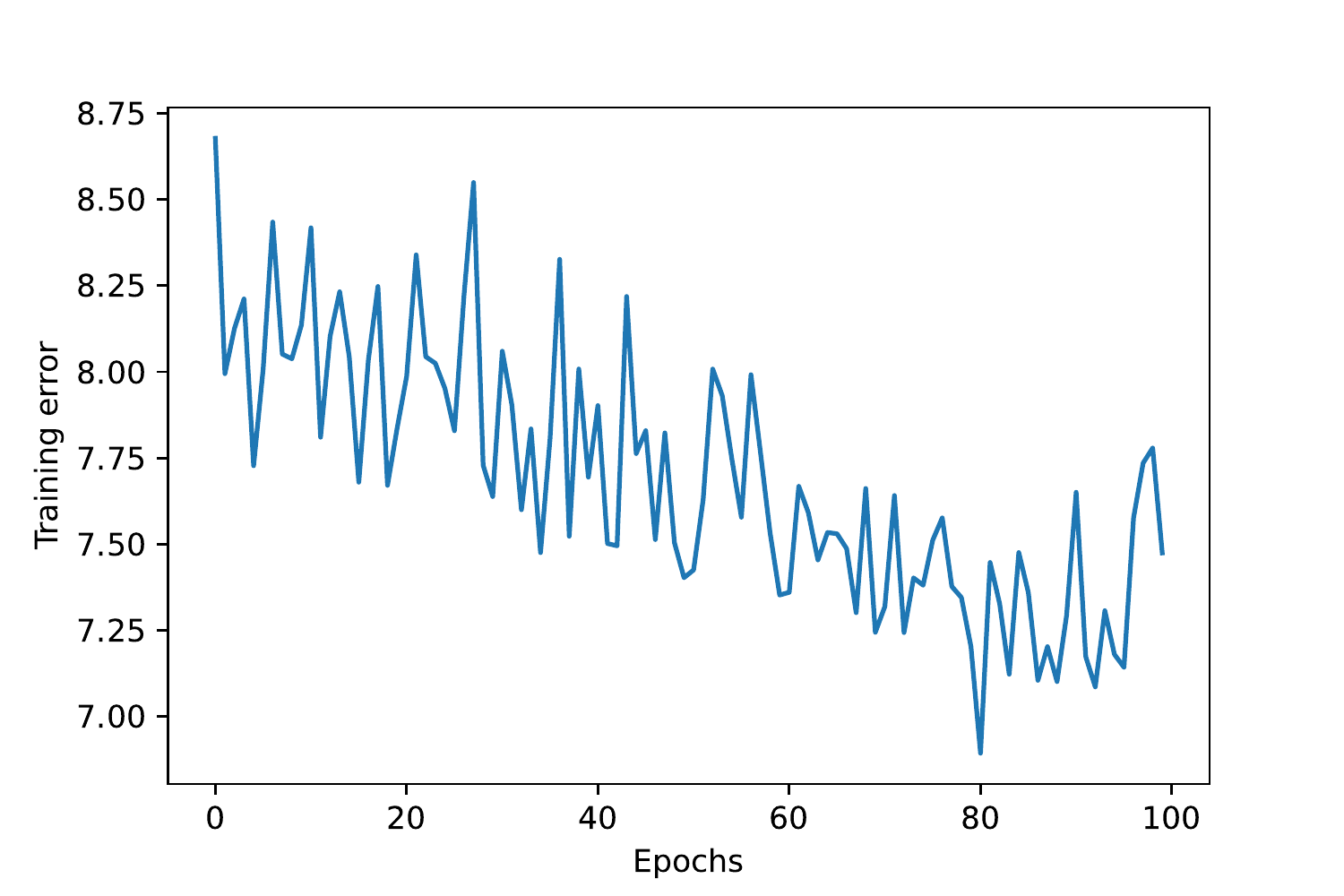}
        \caption{USCA312}
    \end{subfigure}
    \hfill
    \begin{subfigure}{0.30\textwidth}
    \centering
        \includegraphics[width=\textwidth]{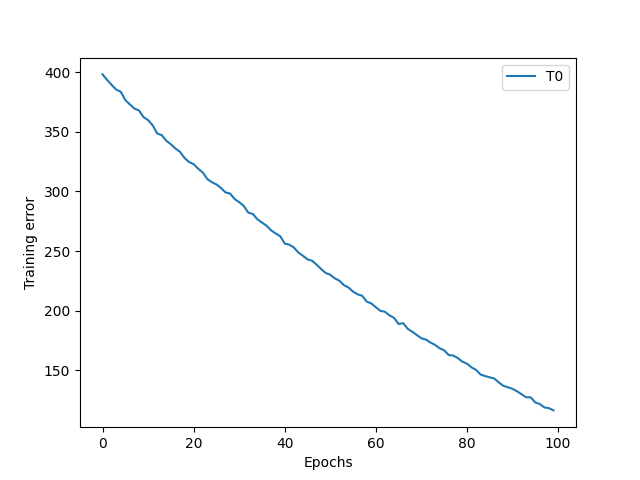}
        \caption{RNAseq}
    \end{subfigure}
    \caption{Training losses from epoch to epoch for real world datasets. }
    \label{fig:convergence}
\end{figure*}

In \Cref{tab:training_time}, we compare the training times required for discrete metric spaces $\mathcal{X}$ with cardinalities ranging from 800 to 1500.  Our ultrametric tree optimization method experiences computational bottleneck in the projection step back to the ultrametric space as we require $O(N^2)$ time per iteration to compute the minimum spanning tree. However, this procedure also allows us to change and optimize over the tree structure from iteration to iteration unlike cTWD and qTWD where the tree structure is fixed. Furthermore, \Cref{fig:convergence} contains the loss per epoch for all real datasets. Notably, in cases such as Twitter, BBCSport, and USAir97, our method demonstrates relatively swift convergence overall even though the time required to project back to ultrametrics at each iteration is significant. However, for Belfast and USCA312, the loss exhibits greater variability from epoch to epoch, implying a more protracted convergence trajectory. Additionally, for the RNAseq data, while the loss steadily decreases, it does not converge within 100 epochs of training which also indicates a more protracted convergence trajectory (similar to the graph datasets). 

\begin{table}[h]
    \centering
    \caption{Comparison of training/construction time (in seconds) for a tree given a discrete metric space $\mathcal{X} \subseteq \R^{2}$ where $|\mathcal{X}|$ varies from 800 to 1500. }
    \begin{tabular}{c|cccc}
         & 800 points & 1000 points & 1500 points \\
         \hline
         Ult. tree training time & 103.00 & 92.79 & 357.86\\
         cTWD construction time & 10.35 & 10.77 & 14.92 \\
         qTWD construction time &   11.72   &   13.55  &  16.90
    \end{tabular}
    
    \label{tab:training_time}
\end{table}

\begin{figure}[t]
\begin{subfigure}{0.45\textwidth}
    \centering
    \includegraphics[width=\textwidth]{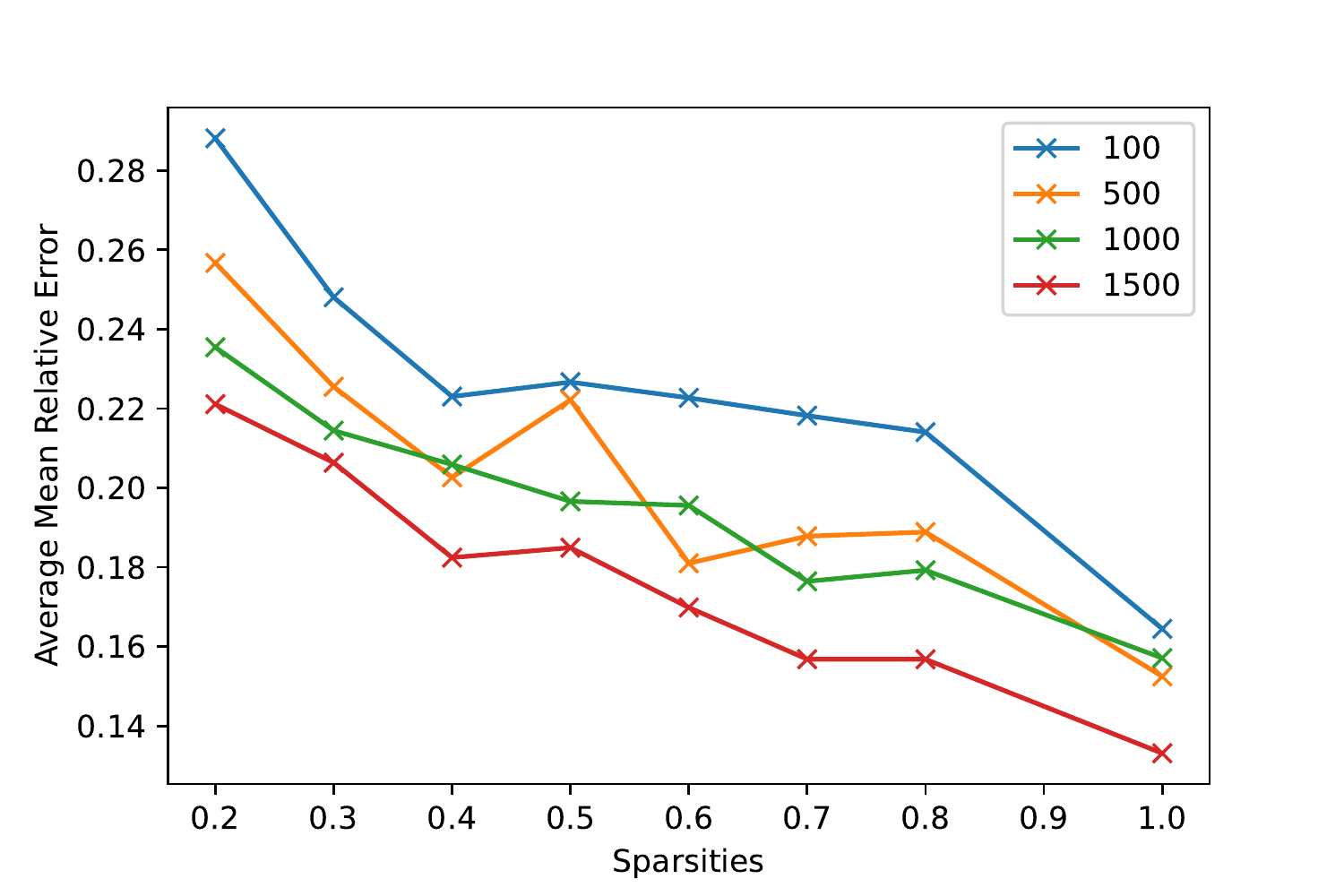}
    \caption{Learned ultrametric where $|\mathcal{X}| \in \{100, 500, 1000, 1500\}$}
\end{subfigure}
\hfill
\begin{subfigure}{0.45\textwidth}
\centering
    \includegraphics[width=\textwidth]{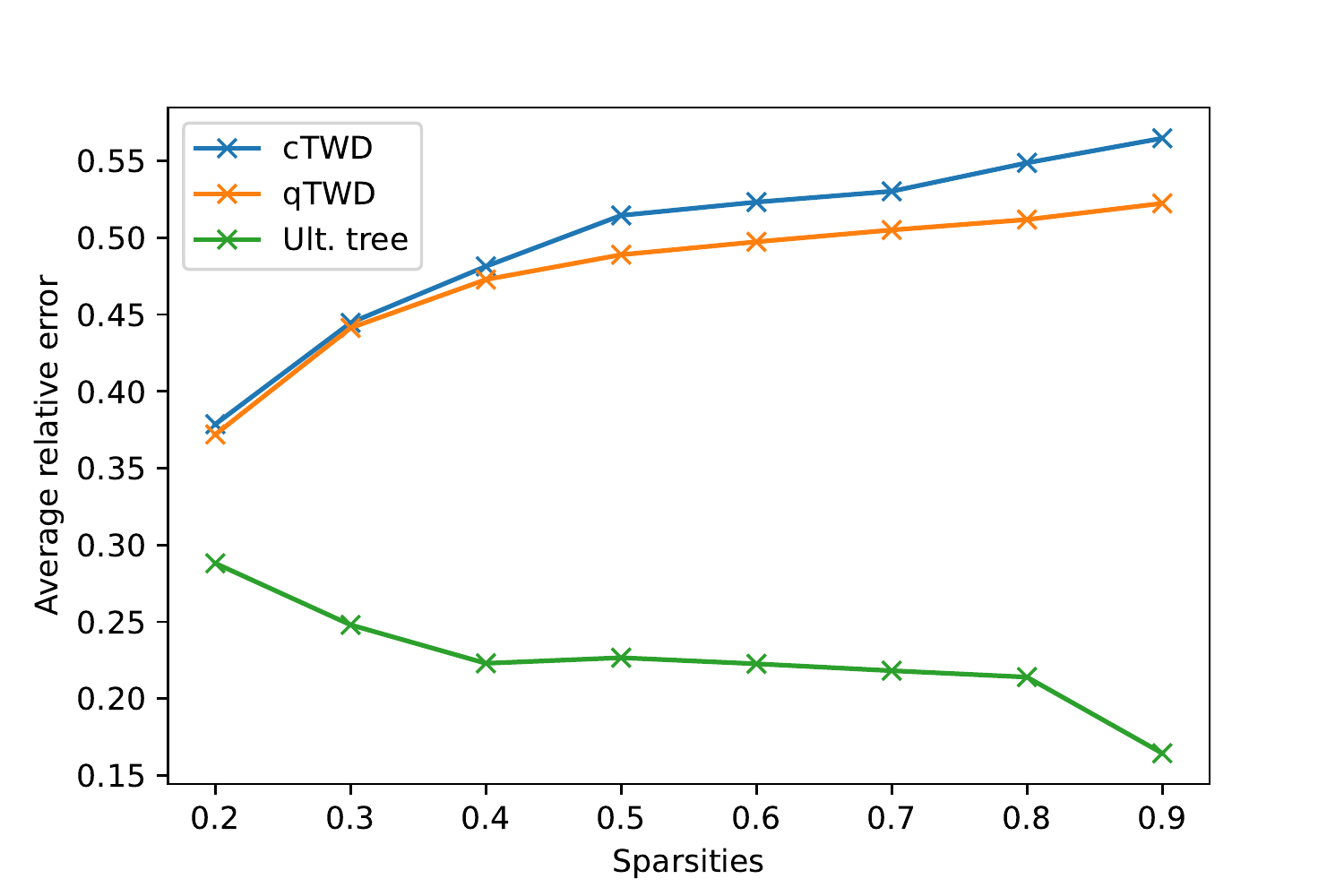}
    \caption{$|\mathcal{X}| = 100$}
\end{subfigure}
\vfill
\begin{subfigure}{0.45\textwidth}
\centering
    \includegraphics[width=\textwidth]{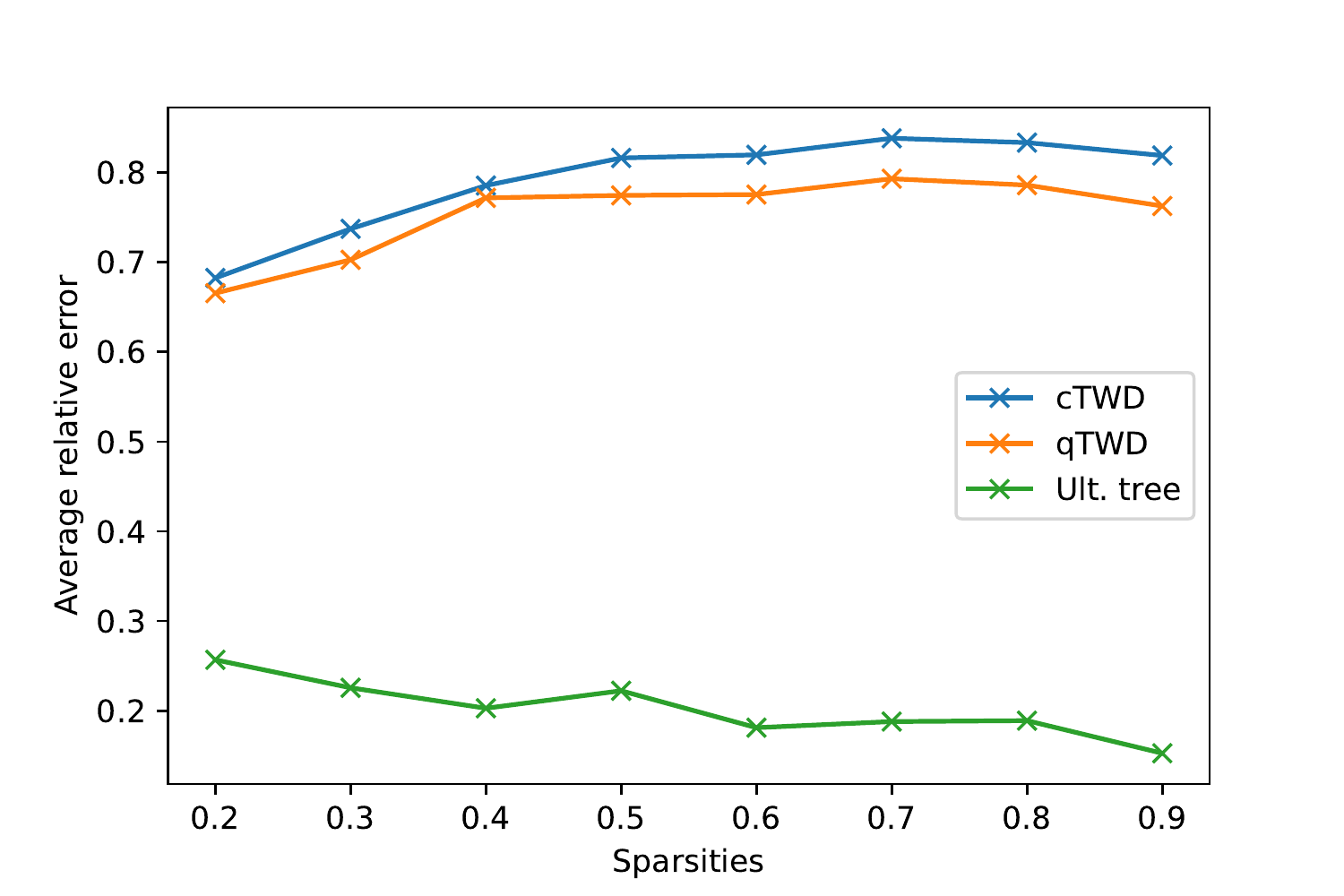}
    \caption{$|\mathcal{X}| = 500$}
\end{subfigure}
\hfill
\begin{subfigure}{0.45\textwidth}
\centering
    \includegraphics[width=\textwidth]{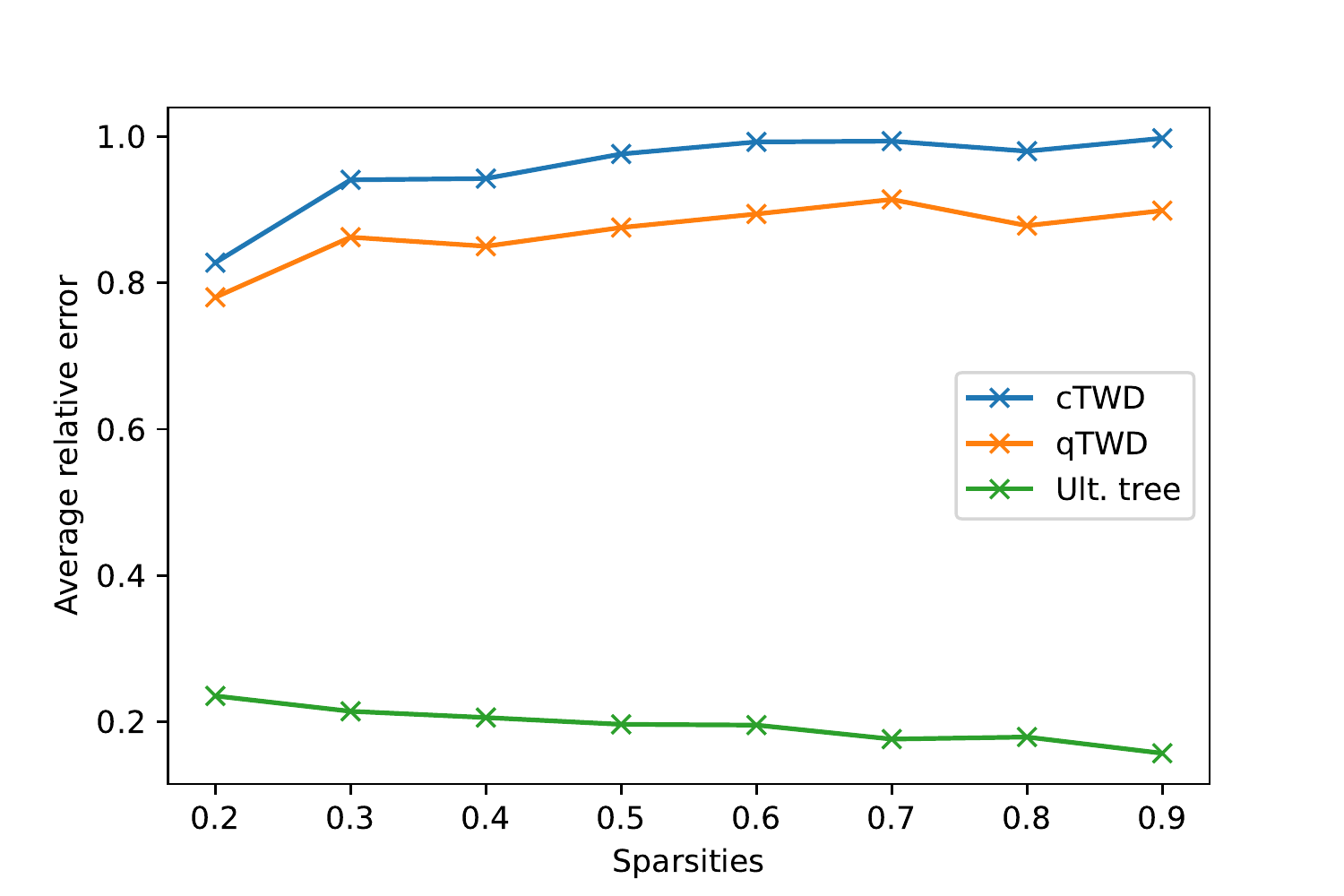}
    \caption{$|\mathcal{X}| = 1000$}
\end{subfigure}
\vfill
\centering
\begin{subfigure}{0.45\textwidth}
\centering
    \includegraphics[width=\textwidth]{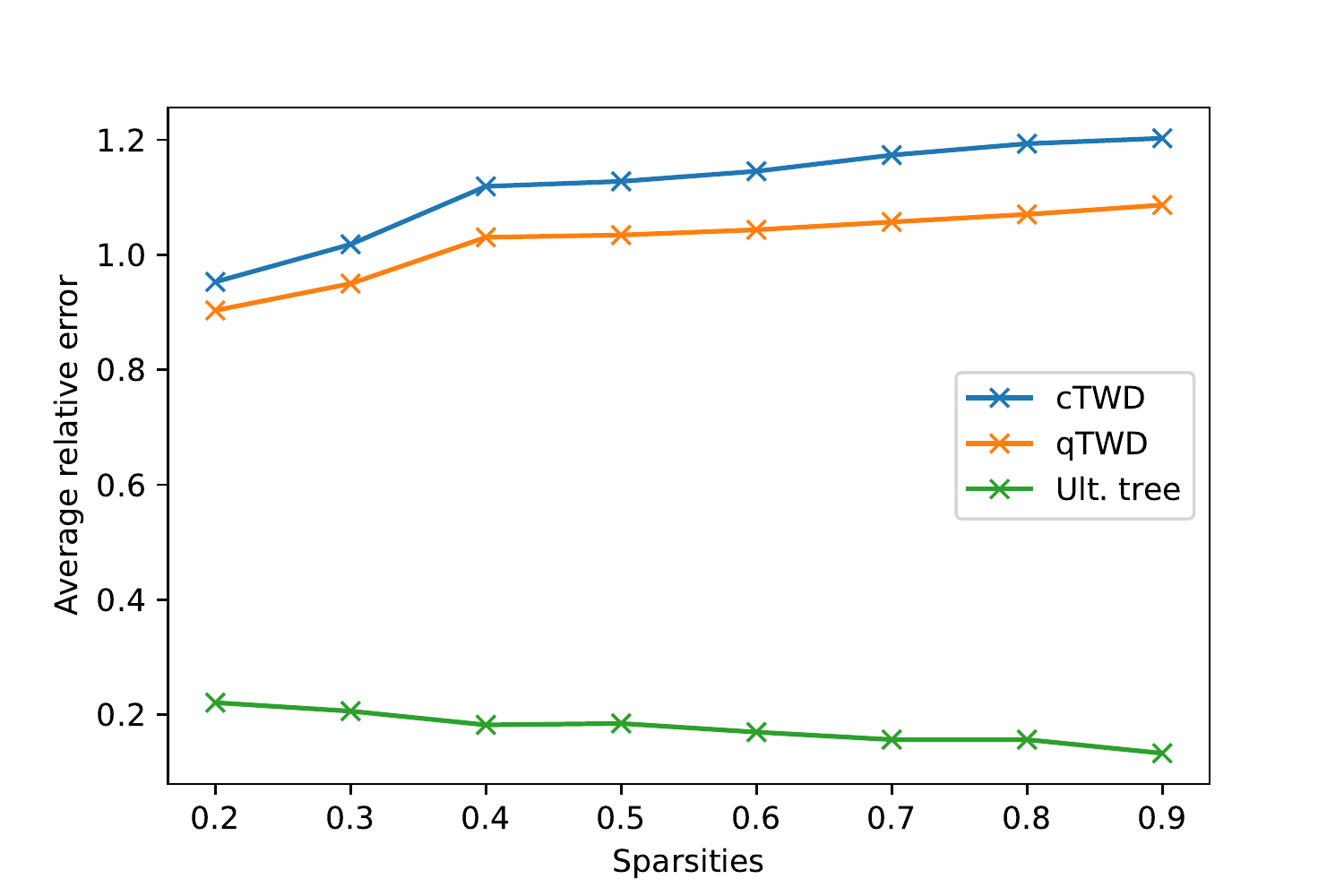}
    \caption{$|\mathcal{X}| = 1500$}
\end{subfigure}
\caption{Changes in the mean relative error as the sparsity of the train and test distributions changes. Plot (a) shows the the changes in average relative error for discrete metric spaces with varying cardinalities as the sparsity of the train and test distributions increase. Plots (b) through (d) compare the average relative error for our learned ultrametric against cTWD and qTWD for discrete metric spaces with varying cardinalities.}
\label{fig:sparsity-plots}
\end{figure} 

\begin{figure*}[t]
    \begin{subfigure}{0.30\textwidth}
    \centering
        \includegraphics[width=\textwidth]{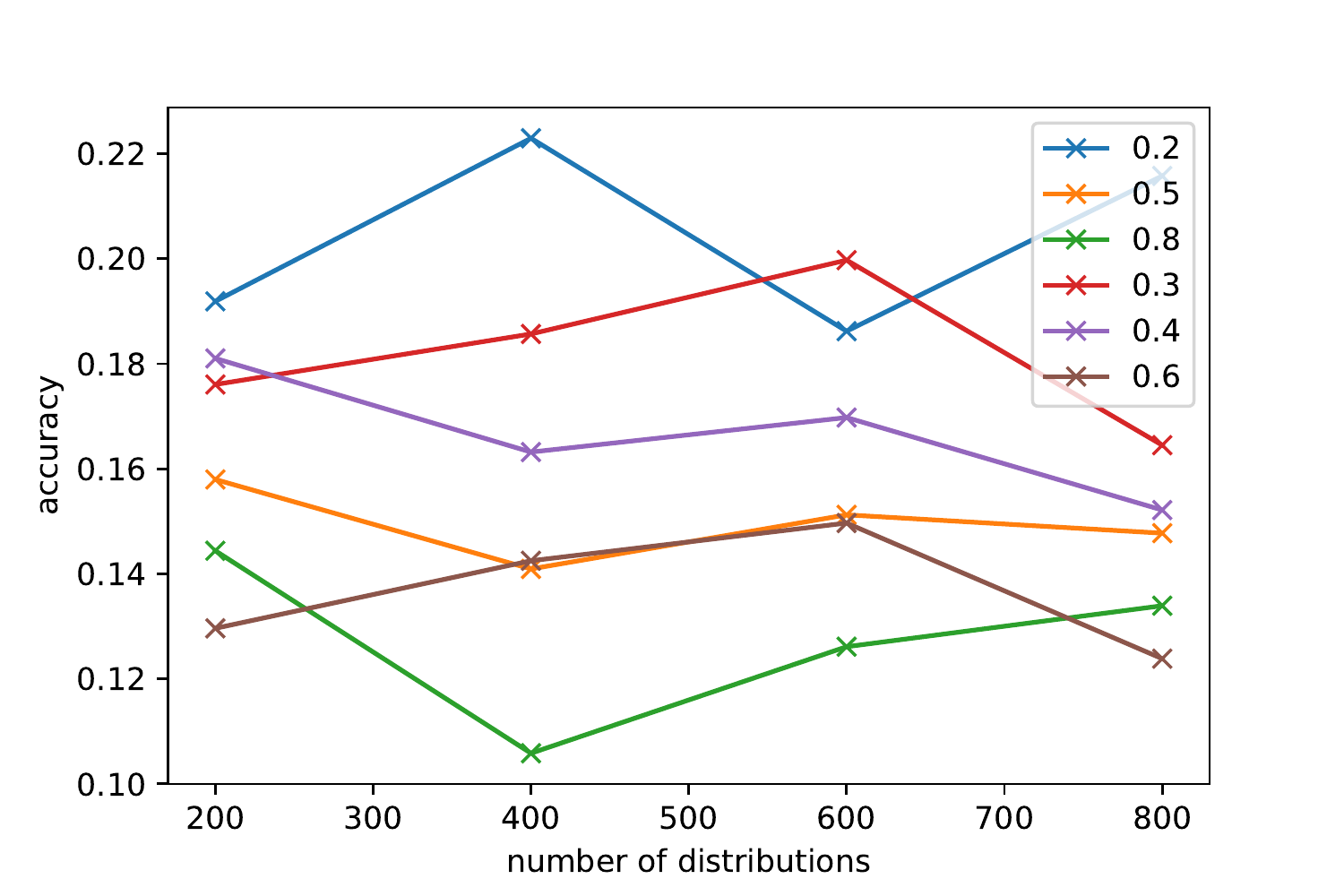}
        \caption{$|\mathcal{X}| = 100$}
    \end{subfigure}
    \hfill
    \begin{subfigure}{0.30\textwidth}
    \centering
        \includegraphics[width=\textwidth]{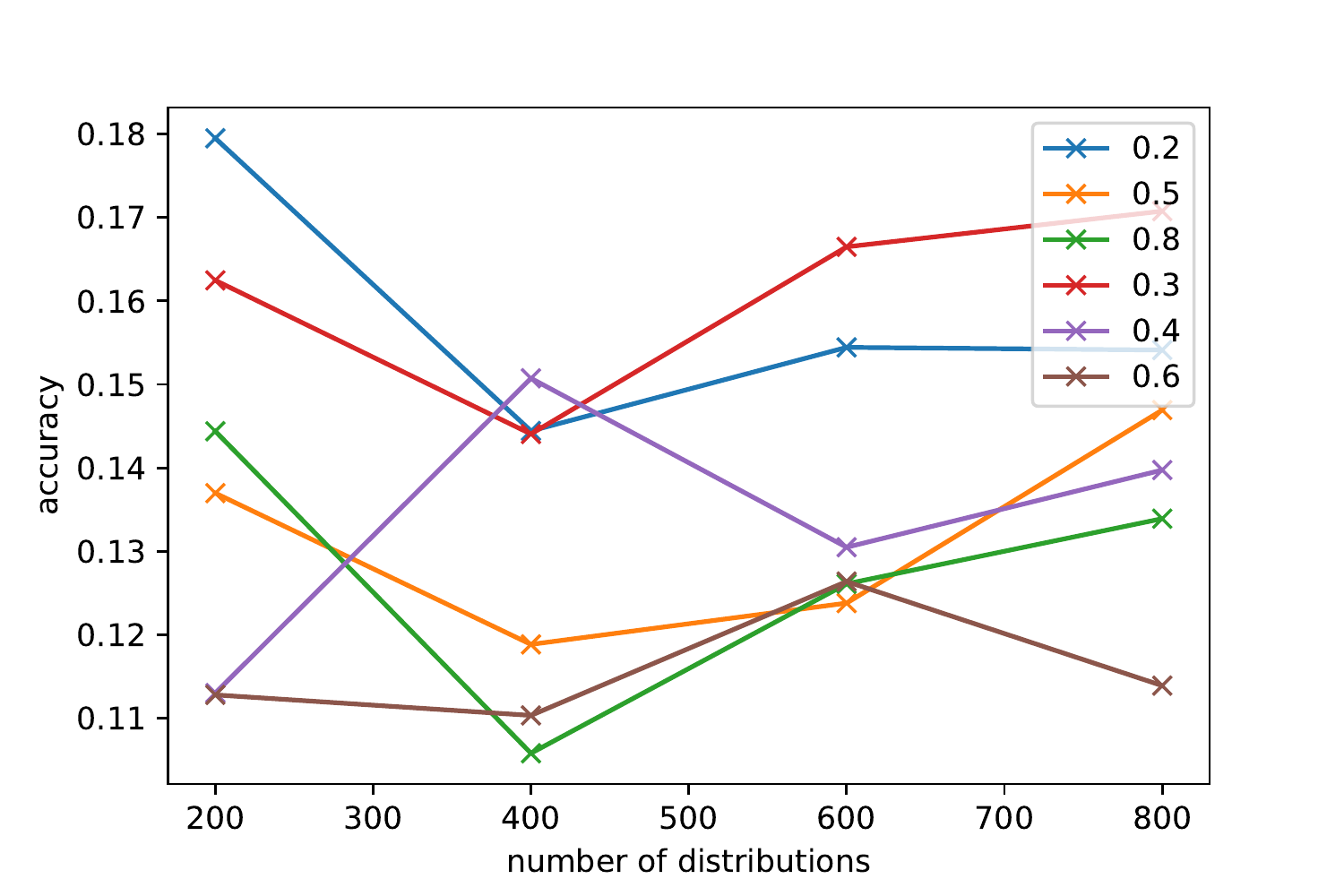}
        \caption{$|\mathcal{X}| = 500$}
    \end{subfigure}
    \hfill
    \begin{subfigure}{0.30\textwidth}
    \centering
        \includegraphics[width=\textwidth]{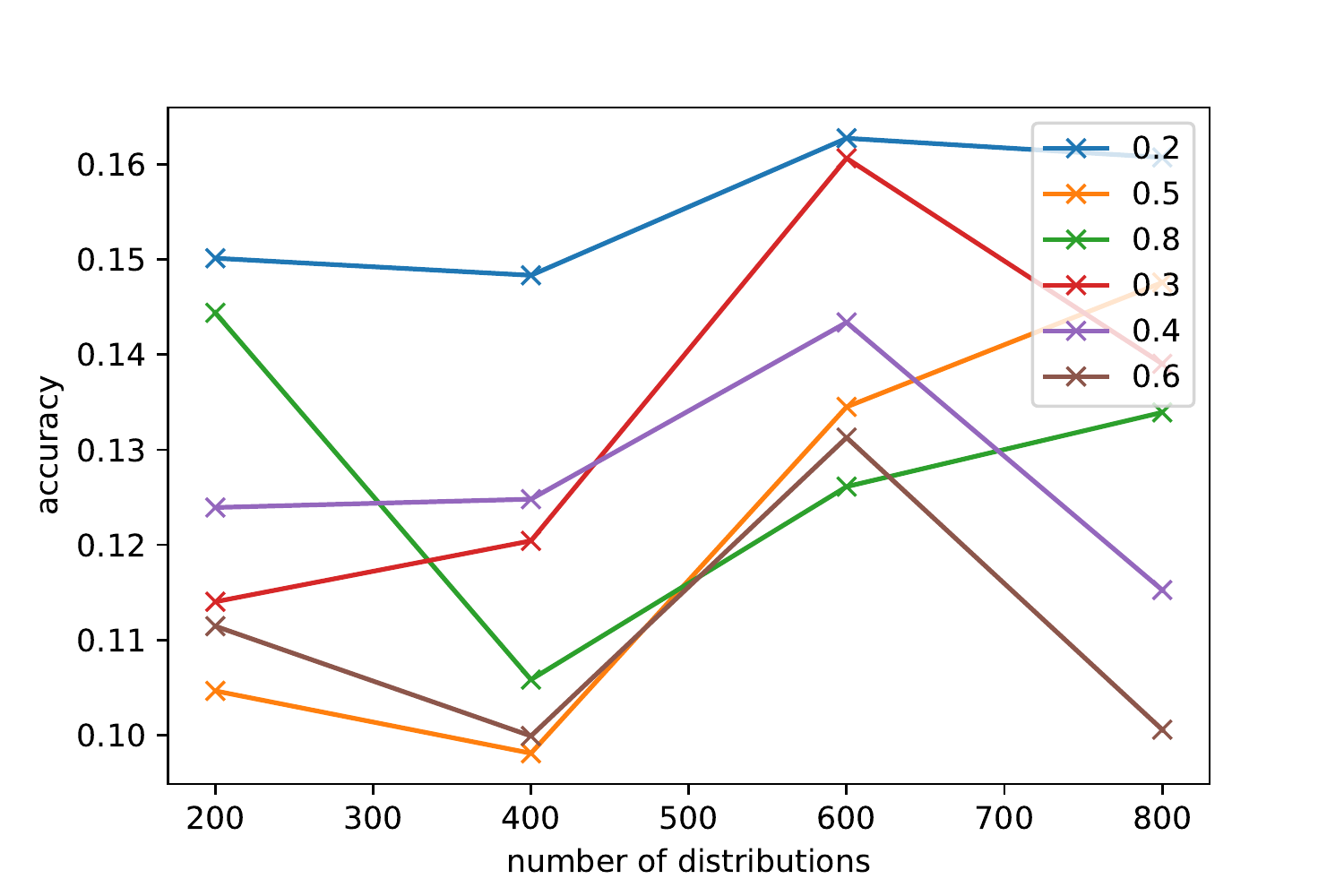}
        \caption{$|\mathcal{X}| = 1000$}
    \end{subfigure}
    \caption{Mean relative error given varying training distribution sizes. }
    \label{fig:vary-distribution-sparsity}
\end{figure*}

\subsection{Additional synthetic datasets}
In addition to synthetic datasets formed by uniformly randomly sampling points on a plane, we also construct a synthetic dataset from multivariate Gaussian distributions in $\R^d$. The results are in \Cref{tab:synthetic_gaussian} and we show that our method outperforms all other methods (including cTWD and qTWD) on an additional synthetic dataset.

\subsection{Effect of sparse distributions}
We examine how the accuracy of our learned ultrametric tree approximation changes as the sparsity of the train and test distributions change. For a randomly generated discrete metric space, $\mathcal{X} \subseteq [-10, 10]^2$, we generate train and test distributions at varying levels of sparsity, $s \in [0, 1]$ by generating random distributions supported on at most $s \cdot |\mathcal{X}|$ elements in $\mathcal{X}$. The results are summarized in \Cref{fig:sparsity-plots} and \Cref{fig:vary-distribution-sparsity}. 
In all cases, as the sparsity of the distribution decreases, the relative error of our learned ultrametric tree approximation decreases. This trend might stem from our increased ability to fine-tune a greater number of nodes for the tree for less sparse distributions, which leads to a more optimal configuration of the tree for such distributions. Additionally, as shown in \Cref{fig:vary-distribution-sparsity}, increasing the size of the training distributions for sparse datasets fails to yield a corresponding reduction in test error for these datasets. Thus, one factor in whether our ultrametric tree optimization method will achieve an accurate approximation of Wasserstein distance is the level of sparsity in for the distributions employed during both training and testing phases.


\end{document}